\documentclass{article}
\usepackage[accepted]{icml2021}

\usepackage{subcaption}
\usepackage{booktabs}       % professional-quality tables

\usepackage{amsfonts,epsfig,graphicx}
\usepackage{amsmath,amssymb,amsthm}

\usepackage{epsf}
\usepackage{epstopdf}
\usepackage{graphics}
\usepackage{amsfonts}
\usepackage{amsmath}
\usepackage{psfrag}
\usepackage{color}
\usepackage{mathtools}
\usepackage[colorinlistoftodos]{todonotes} 
\usepackage{hyperref}
\usepackage{cleveref}

\crefname{equation}{}{}
\newcommand{\lambdamax}[1]{\ensuremath{\lambda_{max}}}

\newcommand{\real}{\ensuremath{\mathbb{R}}}

\newlength{\widebarargwidth}
\newlength{\widebarargheight}
\newlength{\widebarargdepth}

\newtheorem{theo}{Theorem}[section]

\newtheorem{lem}{Lemma}[section]
\newtheorem{prop}{Proposition}[section]
\newtheorem{cor}{Corollary}[section]
\newtheorem{remark}{Remark}[section]

\theoremstyle{definition} 

\newtheorem{nota}{Notation}[section]
\newtheorem{de}{Definition}[section]
\newtheorem{exa}{Example}[section]
\newtheorem{as}{Assumption}[section]
\newtheorem{alg}{Algorithm}[section]

\newcommand{\btheo}{\begin{theo}}
\newcommand{\bde}{\begin{de}}
\newcommand{\ble}{\begin{lem}}
\newcommand{\bpr}{\begin{prop}}
\newcommand{\bno}{\begin{nota}}
\newcommand{\bex}{\begin{exa}}
\newcommand{\bcor}{\begin{cor}}
\newcommand{\spro}{\begin{proof}}
\newcommand{\bas}{\begin{as}}
\newcommand{\balg}{\begin{alg}}
\newcommand{\bremark}{\begin{remark}}

\newcommand{\etheo}{\end{theo}}
\newcommand{\ede}{\end{de}}
\newcommand{\ele}{\end{lem}}
\newcommand{\epr}{\end{prop}}
\newcommand{\eno}{\end{nota}}
\newcommand{\eex}{\end{exa}}
\newcommand{\ecor}{\end{cor}}
\newcommand{\fpro}{\end{proof}}
\newcommand{\eas}{\end{as}}
\newcommand{\ealg}{\end{alg}}
\newcommand{\eremark}{\end{remark}}

\theoremstyle{plain}

\newtheorem{theos}{Theorem}
\newtheorem{props}{Proposition}
\newtheorem{lems}{Lemma}
\newtheorem{cors}{Corollary}
\newtheorem{rems}{Remark}

\theoremstyle{definition}
\newtheorem{exas}{Example}
\newtheorem{algs}{Algorithm}
\newtheorem{asss}{Assumption}
\newtheorem{defns}{Definition}

\newcommand{\btheos}{\begin{theos}}
\newcommand{\etheos}{\end{theos}}
\newcommand{\brems}{\begin{rems}}
\newcommand{\erems}{\end{rems}}
\newcommand{\bprops}{\begin{props}}
\newcommand{\eprops}{\end{props}}
\newcommand{\bdes}{\begin{defns}}
\newcommand{\edes}{\end{defns}}
\newcommand{\blems}{\begin{lems}}
\newcommand{\elems}{\end{lems}}
\newcommand{\bcors}{\begin{cors}}
\newcommand{\ecors}{\end{cors}}
\newcommand{\bexs}{\begin{exas}}
\newcommand{\eexs}{\end{exas}}
\newcommand{\balgs}{\begin{algs}}
\newcommand{\ealgs}{\end{algs}}
\newcommand{\bass}{\begin{asss}}
\newcommand{\eass}{\end{asss}}
\newcommand{\bit}{\begin{itemize}}
\newcommand{\eit}{\end{itemize}}

\newcommand{\data}{{\vec{X}}}
\newcommand{\sample}{\vec{x}}

\newcommand{\weightscalar}{w}
\newcommand{\weight}{\vec{w}}
\newcommand{\weightmat}{\vec{W}}

\newcommand{\dual}{\vec{v}}
\newcommand{\dualmat}{\vec{V}}

\newcommand{\relu}[1]{\left( #1 \right)_+}
\newcommand{\ball}{\mathcal{B}}

\newcommand{\diag}{\mathbf{D}}

\newcommand{\sign}{\text{sign}}
\let\vec\mathbf
\usepackage{hyperref}
\usepackage{url}
\usepackage{tikz}
\usepackage{amssymb}
\usepackage{enumitem}

\makeatletter
\newcommand{\pushright}[1]{\ifmeasuring@#1\else\omit\hfill$\displaystyle#1$\fi\ignorespaces}
\newcommand{\pushleft}[1]{\ifmeasuring@#1\else\omit$\displaystyle#1$\hfill\fi\ignorespaces}
\makeatother

\usepackage[toc,page,header]{appendix}
\usepackage{minitoc}
\renewcommand{\qed}{$\blacksquare$}
\usepackage{bbm}

\icmltitlerunning{Global Optimality Beyond Two Layers: Training Deep ReLU Networks via Convex Programs}

\begin{document}
\doparttoc % Tell to minitoc to generate a toc for the parts
\faketableofcontents % Run a fake tableofcontents command for the partocs

\twocolumn[
\icmltitle{Global Optimality Beyond Two Layers: Training Deep ReLU Networks via Convex Programs}




\begin{icmlauthorlist}
\icmlauthor{Tolga Ergen}{to}
\icmlauthor{Mert Pilanci}{to}

\end{icmlauthorlist}

\icmlaffiliation{to}{Department of Electrical Engineering, Stanford University, CA, USA}

\icmlcorrespondingauthor{Tolga Ergen}{ergen@stanford.edu}
\icmlcorrespondingauthor{Mert Pilanci}{pilanci@stanford.edu}

\icmlkeywords{Machine Learning, ICML}

\vskip 0.3in
]



\printAffiliationsAndNotice{} % otherwise use the standard text.

%%%%%%%%% ABSTRACT
\begin{abstract}
Understanding the fundamental mechanism behind the success of deep neural networks is one of the key challenges in the modern machine learning literature. Despite numerous attempts, a solid theoretical analysis is yet to be developed. In this paper, we develop a novel unified framework to reveal a hidden regularization mechanism through the lens of convex optimization. We first show that the training of multiple three-layer ReLU sub-networks with weight decay regularization can be equivalently cast as a convex optimization problem in a higher dimensional space, where sparsity is enforced via a group $\ell_1$-norm regularization. Consequently, ReLU networks can be interpreted as high dimensional feature selection methods. More importantly, we then prove that the equivalent convex problem can be globally optimized by a standard convex optimization solver with a polynomial-time complexity with respect to the number of samples and data dimension when the width of the network is fixed. Finally, we numerically validate our theoretical results via experiments involving both synthetic and real datasets.
\end{abstract}

%%%%%%%%%%%%%%%%%%%%%%%%%%%%%%% new section %%%%%%%%%%%%%%%%%%%%%%%%%%%%%%%%%%%%%%%%%
\section{Introduction}
Deep neural networks have been extensively studied in machine learning, natural language processing, computer vision, and many other fields. Even though deep neural networks have provided dramatic improvements over conventional learning algorithms \cite{goodfellow2016deep}, fundamental mathematical mechanisms behind this success remain elusive. To this end, in this paper, we investigate the implicit mechanisms behind deep neural networks by leveraging tools available in convex optimization theory.

\begin{figure}[h]
        \centering
            \centering
            \includegraphics[width=0.5\textwidth,height=0.4\textwidth]{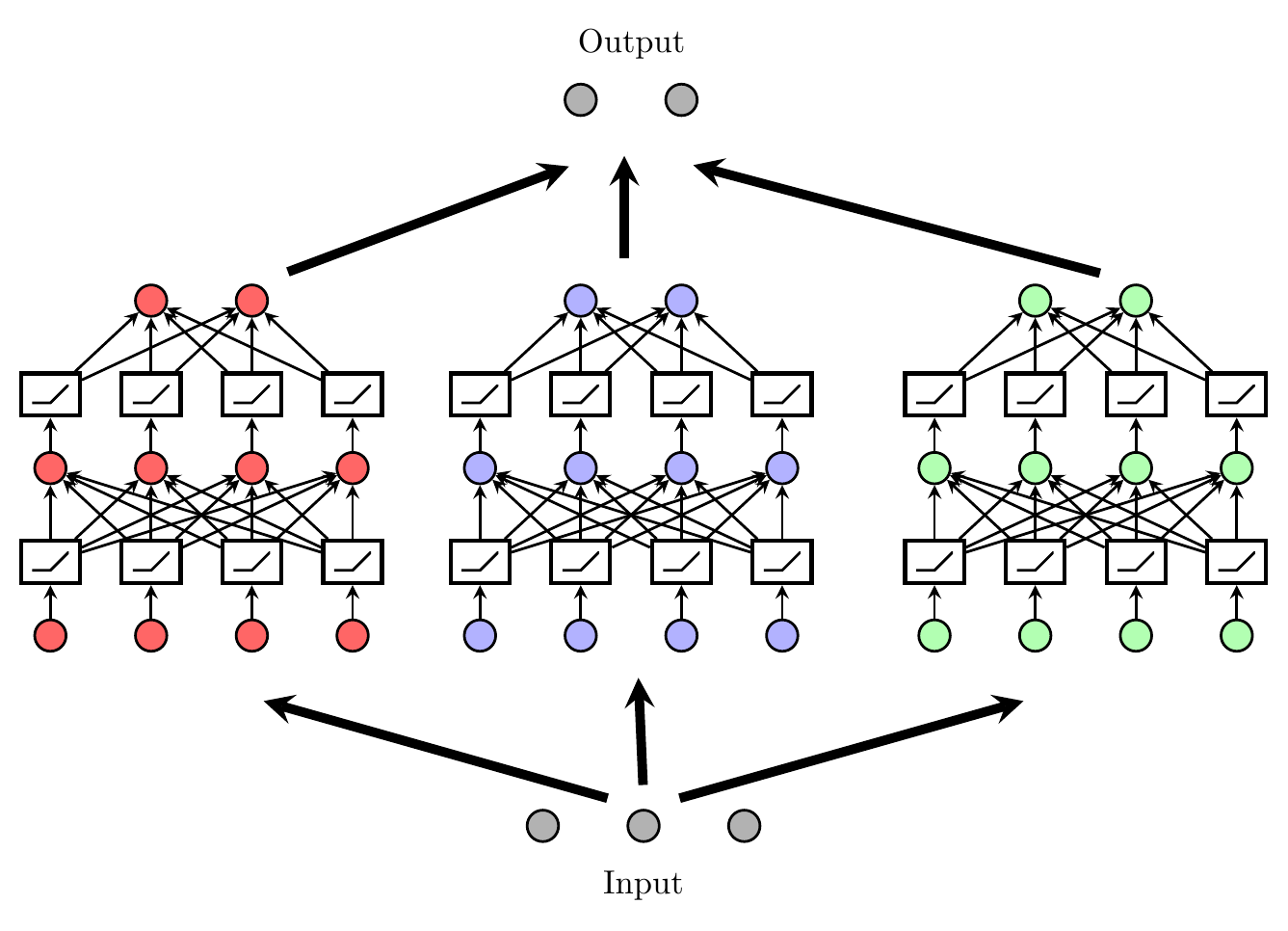}
            \caption{An architecture with three sub-networks, i.e., $K=3$, where each is a three-layer ReLU network.}\label{fig:subnetworks}%\vskip -0.2in
    \end{figure}

\begin{figure*}[t]
        \centering
        \begin{subfigure}[b]{0.4\textwidth}
            \centering
            \includegraphics[width=\textwidth,height=0.7\textwidth]{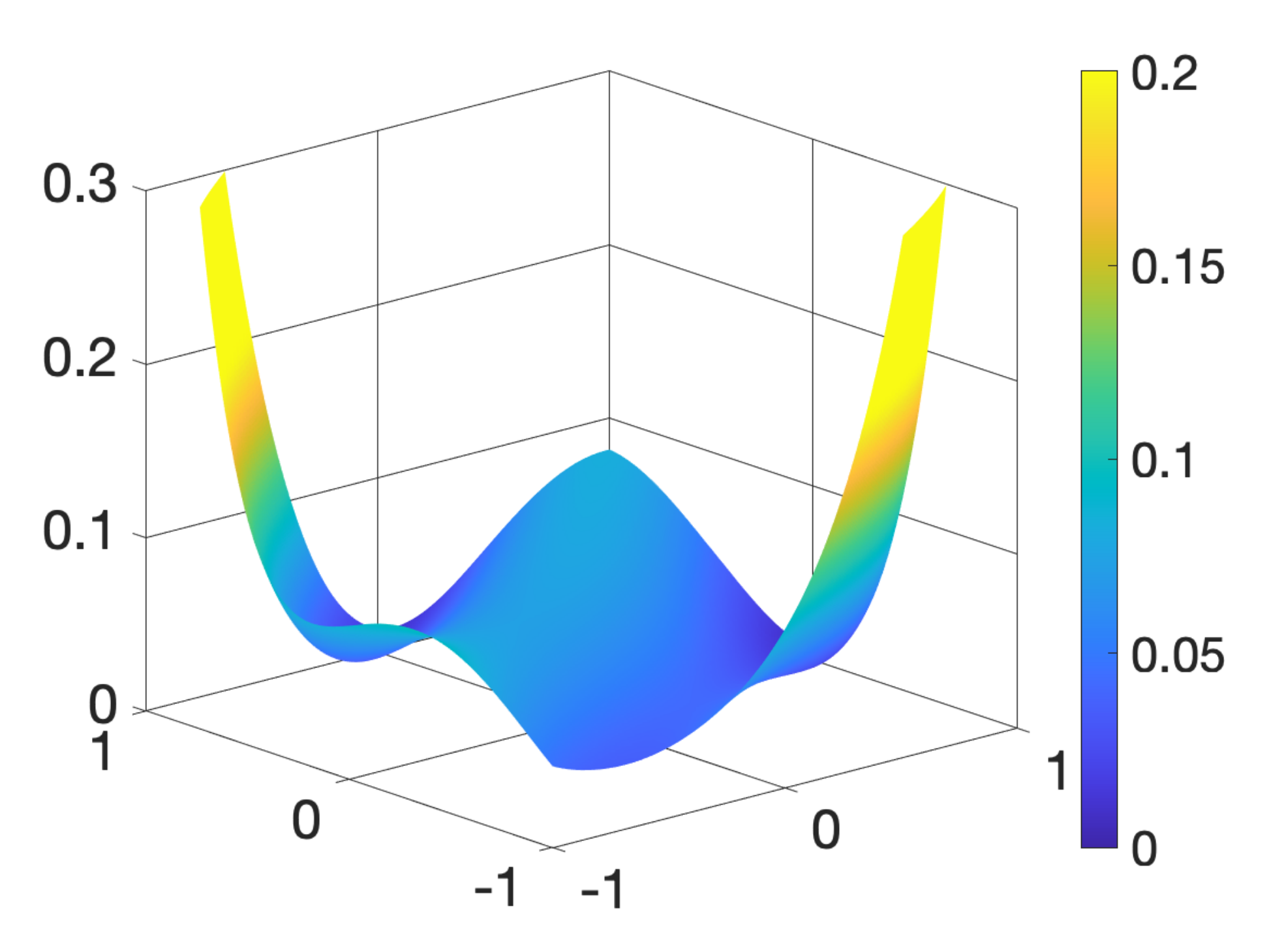}
            \caption{$K=1$}
        \end{subfigure}
        \hfill
        \begin{subfigure}[b]{0.4\textwidth}  
            \centering 
            \includegraphics[width=\textwidth,height=0.7\textwidth]{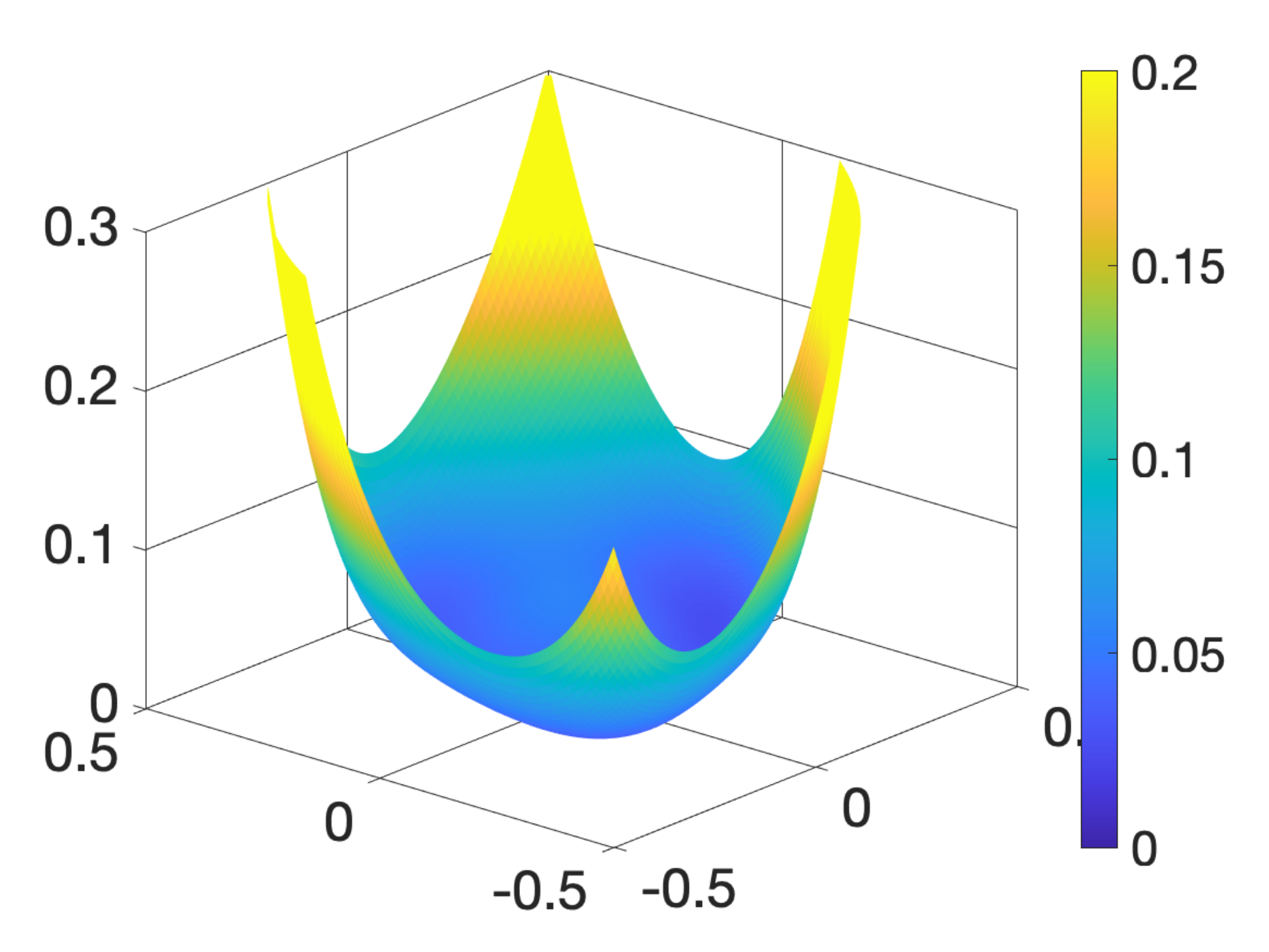}
            \caption{$K=50$}
            \end{subfigure}
	\caption{The loss landscape of an architecture consisting of $K$ three-layer ReLU networks trained on a synthetic dataset, where the number of samples, features, and neurons are chosen as $(n,d,m_1)=(100,10,10)$, respectively. For the dataset, we first randomly generate a data matrix by sampling a multivariate standard normal distribution and then obtain the corresponding label vector by feeding the data matrix into a randomly initialized teacher network with $m_1=10$.  }\label{fig:landscape_synthetic}
	%\vskip -0.2in
    \end{figure*}

%%%%%%%%%%%%%%%%%%%%%%%%%%%%%%% new subsection %%%%%%%%%%%%%%%%%%%%%%%%%%%%%%%%%%%%%%%%%
\subsection{Related Work}
The problem of learning the parameters of a deep neural network is the subject of 
many studies due to its highly non-convex and nonlinear nature. Despite their complex structure, deep networks are usually trained by simple local search algorithms such as Gradient Descent (GD) to achieve a globally optimal set of parameters \cite{brutzkus_localminima}. However, it has been shown that these algorithms might also converge a local minimum in pathological cases \cite{shalev2017failures,goodfellow2016deep,ergen2019shallow}. In addition to this, \cite{ge2017onehidden,shamir2018spurious} proved that Stochastic GD (SGD) is likely to get stuck at a local minimum when the number of parameters is small. Furthermore, \cite{anandkumar2016saddle} reported that complicated saddle points do exist in the optimization landscape of deep networks. Since such points might be hard to escape for local search algorithms, training deep neural networks is a computationally challenging optimization problem \cite{dasgupta1995complexity,blum1989threenode,bartlett1999hardness}.

In order to alleviate these training issues, a line of research focused on designing new architectures that enjoy a well behaved optimization landscape by leveraging overparameterization \cite{brutzkus_overparameterized_linear,du2018overparameterized,arora2018overparameterization,neyshabur2018overparameterization} and combining multiple neural network architectures, termed as sub-networks, in parallel \cite{iandola2016squeezenet,szegedy2017inception,chollet2017xception,xie2017aggregatedresidual,zagoruyko2016wideresidual,veit2016ensembleresidual} as illustrated in Figure \ref{fig:subnetworks}. Such studies empirically proved that increasing the number of sub-networks yields less complicated optimization landscapes so that GD generally converges to a global minimum. These empirical observations are due to the fact that increasing the number of sub-networks yields a less non-convex optimization landscape. To support this claim, we also provide an experiment in Figure \ref{fig:landscape_synthetic}, where increasing the number of sub-network clearly promotes convexity of the loss landscape. In addition to better training performance, neural network architectures with multiple sub-networks also enjoy a remarkable generalization performance so that various such architectures have been introduced to achieve state-of-the-art performance in practice, especially for image classification tasks. As an example, SqueezeNet \cite{iandola2016squeezenet}, Inception \cite{szegedy2017inception}, Xception \cite{chollet2017xception}, and ResNext \cite{xie2017aggregatedresidual} are combinations of multiple networks and achieved notable improvements in practice.

%%%%%%%%%%%%%%%%%%%%%%%%%%%%%%% new subsection %%%%%%%%%%%%%%%%%%%%%%%%%%%%%%%%%%%%%%%%%
\subsection{Our Contributions}
Our contributions can be summarized as follows:
\begin{itemize}[leftmargin=*]
\item We introduce an exact analytical framework based on convex duality to characterize the optimal solutions to regularized deep ReLU network training problems. As a corollary, we provide interpretations for the convergence of local search algorithms such as SGD and the loss landscape of these training problems. We also numerically verify these interpretations via experiments involving both synthetic and real benchmark datasets.

\item  We show that the training problem of an architecture with multiple ReLU sub-networks can be equivalently stated as a convex optimization problem. More importantly, we prove that the equivalent convex problem can be globally optimized in polynomial-time using standard convex optimization solvers. Therefore, we prove the polynomial-time trainability of regularized ReLU networks with multiple nonlinear layers, which generalizes the recent two-layer results in \cite{pilanci2020convex,ergen2020aistats,ergen2020journal} to a much broader class of neural network architectures.

\item Our analysis also reveals an implicit regularization structure behind the non-convex ReLU network training problems. In particular, we show that this implicit regularization is a group $\ell_1$-norm regularization, which encourages sparsity for the equivalent convex problem in a high dimensional space. We further prove that there is a direct mapping between the original non-convex and the equivalent convex training problems so that sparsity for the convex problem implies a smaller number of sub-networks for the original non-convex problem.

\item Unlike the previous studies, our results hold for arbitrary convex loss functions including squared, cross entropy, and hinge loss, and common regularization methods, e.g., weight decay. 

\end{itemize}

%%%%%%%%%%%%%%%%%%%%%%%%%%%%%%% new subsection %%%%%%%%%%%%%%%%%%%%%%%%%%%%%%%%%%%%%%%%%%
\subsection{Notation
}
We denote matrices and vectors as uppercase and lowercase bold letters, respectively. We use $\vec{I}_k$ to denote the identity matrix of size $ k \times k$ and $\vec{0}$ (or $\vec{1}$) to denote a vector/matrix of zeros (or ones) with appropriate sizes. We denote the set of integers from $1$ to $n$ as $[n]$. In addition, $\|\cdot\|_2$ and $\|\cdot \|_{F}$ denotes the Euclidean and Frobenius norms, respectively. Furthermore, we define the unit $\ell_p$-ball as $\ball_p:=\{\vec{u} \in \mathbb{R}^d\, :\,\|\vec{u}\|_p\le 1\}$. We also use $\mathbf{1}[x\geq0]$ and $\relu{x}=\max\{x,0\}$ to denote the elementwise 0-1 valued indicator and ReLU, respectively.

%%%%%%%%%%%%%%%%%%%%%%%%%%%%%%% new subsection %%%%%%%%%%%%%%%%%%%%%%%%%%%%%%%%%%%%%%%%%
\subsection{Overview of Our Results}
In this paper, we consider an architecture with $K$ sub-networks each of which is an $L$-layer ReLU network with layer weights $\weightmat_{lk} \in \mathbb{R}^{m_{l-1} \times m_l}$, $\forall l \in [L]$, where $m_0=d$, $m_L=1$\footnote{We consider scalar output networks for presentation simplicity,  however, all our derivations can be extended to vector output networks as shown in Section \ref{sec:supp_vectorout} of the supplementary file. }, and $m_l$ denotes the number of neurons in the $l^{th}$ hidden layer. Given an input data matrix $\data \in \mathbb{R}^{n \times d}$ and the corresponding label vector $\vec{y} \in \mathbb{R}^n$, the regularized network training problem can be formulated as follows
\begin{align}
    \min_{\theta \in \Theta} \mathcal{L} \left( \sum_{k=1}^K f_{\theta,k}(\data),\vec{y} \right)+\beta  \sum_{k=1}^K\mathcal{R}_k(\theta) \,, \label{eq:problem_statement}
\end{align}
where $\Theta$ is the parameter space, $\mathcal{L}(\cdot,\cdot)$ is an arbitrary convex loss function, $\mathcal{R}_k(\cdot)$ is the regularization function for the layer weights in the $k^{th}$ sub-network, and $\beta>0$ is a regularization parameter. In addition to this, we compactly define the set of network parameters as $\theta:=\{\{\weightmat_{lk}\}_{l=1}^L\}_{k=1}^K$, $\theta \in \Theta$ and the output of each sub-network as 
\begin{align}\label{eq:sub-network}
    f_{\theta,k}(\data):= \relu{\relu{\data \weightmat_{1k}} \ldots \weight_{(L-1)k} }\weightscalar_{Lk}.
\end{align}
\begin{rems}
Notice that the function parallel architectures include a wide range of neural network architectures in practice. As an example, ResNets \cite{resnet} are a special case of this architecture. We first note that residual blocks are applied after ReLU in practice so that the input to each block has nonnegative entries. Hence, for this special case, we assume $\data \in \mathbb{R}_+^{n \times d}$. Let us consider a four-layer architecture with $K=2$, $\weightmat_{11}=\weightmat_1$, $\weightmat_{21}=\weightmat_2$, $\weightmat_{12}=\weightmat_{22}=\vec{I}_d$, $\weight_{31}=\weight_{32}=\weight_3$, and $\weightscalar_{41}=\weightscalar_{42}=\weightscalar_4$ then
\begin{align*}
    f_{\theta,k}(\data)&=\sum_{k=1}^2 \relu{\relu{\relu{\data \weightmat_{1k}} \weightmat_{2k}} \weight_{3k} }\weightscalar_{4k}\\
    &=\relu{\relu{\relu{\data \weightmat_{1}} \weightmat_{2}} \weight_{3} }\weightscalar_{4}+\relu{\data \weight_{3} }\weightscalar_{4}
\end{align*}
which corresponds to a shallow ResNet as depicted in Figure 1 of \cite{veit2016ensembleresidual}.
\end{rems}
 Throughout the paper, we consider the conventional regression framework with weight decay regularization and squared loss, i.e., $\mathcal{L}( f_{\theta}(\data),\vec{y})=\|f_{\theta}(\data)-\vec{y}\|_2^2$ and $\mathcal{R}(\theta)=\frac{1}{2} \|\theta\|_2^2$. However, our derivations also hold for arbitrary convex loss functions including hinge loss and cross entropy and vector outputs as proven in the supplementary file. Thus, we consider the following optimization problem
\begin{align}
    P^*:=& \min_{\theta \in \Theta} \mathcal{L}\left( \sum_{k=1}^K f_{\theta,k}(\data),\vec{y}\right)+ \frac{\beta}{2} \sum_{k=1}^K\sum_{l=L-1}^L \|\weightmat_{lk}\|_F^2 , \label{eq:problem_def_overview}
\end{align}
where $\Theta:=\{\theta: \|\weightmat_{lk}\|_F \leq 1, \forall l \in [L-2], \forall k \in [K]\}$ without loss of generality.

%%%%%%%%% remark %%%%%%
\begin{rems}\label{remark:regularization}
In \eqref{eq:problem_def_overview}, we impose unit Frobenius norm constraints on the first $L-2$ layer weights and regularize only the last two layers. Although this might appear to limit the effectiveness of the regularization on the network output, in Lemma \ref{lemma:scaling_deep_overview}, we show that as long as the last two layers' weights of each sub-network are regularized, the remaining layer weights do not change the structure of the regularization. They only contribute to the ratio between the training error and regularization term. Therefore, one can undo this by simply tuning $\beta$ (see Section \ref{sec:proof_remark2} of the supplementary file for details).
\end{rems}

Next, we introduce a rescaling technique to equivalently state the problem in \eqref{eq:problem_def_overview} as an $\ell_1$-norm minimization problem, which is critical for strong duality to hold.

%%%%%%% lemma %%%%%%%%%%
\begin{lems}\label{lemma:scaling_deep_overview}
The following problems are equivalent \footnote{All the proofs are presented in the supplementary file.}:
\begin{align*}
\begin{split} \min_{\theta \in \Theta}  \mathcal{L} \left( \sum_{k=1}^K f_{\theta,k}(\data),\vec{y} \right)+ \frac{\beta}{2}\sum_{k=1}^K \sum_{l=L-1}^L \|\weightmat_{lk}\|_F^2
\end{split}\\
    = \begin{split}
         &\min_{{\theta \in \Theta_p} } \mathcal{L}\left(\sum_{k=1}^K f_{\theta,k}(\data),\vec{y} \right)+ \beta \sum_{k=1}^K |\weightscalar_{Lk}|
\end{split},
\end{align*}
where $\Theta_p:=\{\theta \in \Theta: \|\weightmat_{lk}\|_F\leq 1, \forall l \in [L-2],\;\|\weight_{(L-1)k}\|_2\leq 1, \forall k \in [K]\}$.
\end{lems}

Using Lemma \ref{lemma:scaling_deep_overview}, we first take the dual of $\ell_1$ equivalent of \eqref{eq:problem_def_overview} with respect to the output weights $\weightscalar_{Lk}$ and then change the order of min-max to achieve the following dual problem, which provides a lower bound for the primal problem \eqref{eq:problem_def_overview}\footnote{We present the details in Section \ref{sec:dual_derivation} of the supplementary file.}
\begin{align}\label{eq:dual_overview}
        &P^*\geq D^* :=\max_{\vec{\dual}} - \mathcal{L}^*(\dual)   \\ \nonumber
        &\text{s.t. } \max_{\theta \in \Theta_p} \left\vert \dual^T \relu{\relu{\data \weightmat_{1}}\ldots \weight_{(L-1)}} \right\vert\leq \beta,
\end{align}
where $\mathcal{L}^*$ is the Fenchel conjugate function defined as \cite{boyd_convex}
\begin{align*}
\mathcal{L}^*(\dual) := \max_{\vec{z}} \vec{z}^T \dual - \mathcal{L}(\vec{z},\vec{y})\,.
\end{align*}
Using this dual characterization, we first find a set of hidden layer weights via the optimality conditions (i.e. active dual constraints). We then prove the optimality of these weights via strong duality, i.e., $P^*=D^*$.

%%%%%%%%%%%%%%%%%%%%%%%%%%%%%%% new subsection %%%%%%%%%%%%%%%%%%%%%%%%%%%%%%%%%%%%%%%%%
\subsection{Prior Work \cite{zhang2019multibranch,haeffele2017global,pilanci2020convex}}
Here, we further clarify contributions and limitations of some recent studies \cite{zhang2019multibranch,haeffele2017global,pilanci2020convex} that focus on the training problem of architectures with sub-networks through the lens of convex optimization theory. \cite{haeffele2017global} particularly analyzed the characteristics of the local minima of the regularized training objective in \eqref{eq:problem_statement}. However, the results are valid under several impractical assumptions. As an example, they require all local minima of \eqref{eq:problem_statement} to be rank-deficient. Additionally, they assume that the objective function \eqref{eq:problem_statement} is twice differentiable, which is not the case for non-smooth problems, e.g., training problems with ReLU activation. Furthermore, they require $K$ to be too large to be of practical use. Finally, their proof techniques depend on finding a local descent direction of a non-convex training problem, which might be NP hard in general. Therefore, even though this study provided valuable insights for future research, it is far from explaining observations in practical scenarios.

In addition to \cite{haeffele2017global}, \cite{zhang2019multibranch} provided some results on strong duality. They particularly showed that the primal-dual gap diminishes as the number of sub-networks $K$ increases. Although this study is an important step to understand deep networks through convex duality, it does not include any solid results for finite-size networks. Moreover, their results require strict assumptions: 1) the analysis only works for hinge loss and linear networks; 2) the analysis requires the data matrix to be included in the regularization term $\mathcal{R}(\theta)$, thus, it is not valid for commonly used regularizations such as weight decay in \eqref{eq:problem_def_overview}; 3) they require assumptions on the regularization parameter $\beta$.

Another closely related work \cite{pilanci2020convex} studied convex optimization for ReLU networks and followed by a series of papers \cite{sahiner2021vectoroutput,ergen2020cnn,ergen2021bn,vikul2021generative}. Particularly, the authors introduced an exact convex formulation to train two-layer ReLU networks in polynomial-time for training data $\data \in \mathbb{R}^{n \times d}$ of constant rank, where the network output is $f_{\theta}(\data):= \sum_{j=1}^m \relu{\data \vec{u}_j} \alpha_j$ given the hidden layer weights $\vec{u}_j \in \mathbb{R}^{d}$ and the output layer weights $\alpha_j \in \mathbb{R}$. However, their analysis does not extend to deep architectures with more than one ReLU layer. The reason for this limitation is that the composition of multiple ReLU layers is a significantly more challenging optimization problem. Moreover, in such a case, the complexity becomes exponential-time due to the complex and combinatorial behavior of multiple ReLU layers.

%%%%%%%%%%%%%%%%%%%%%%%%%%%%%%% new section %%%%%%%%%%%%%%%%%%%%%%%%%%%%%%%%%%%%%%%%%
\section{Architectures with Three-layer ReLU Sub-networks}

Here, we consider $K$ three-layer ReLU sub-networks, i.e., $L=3$, trained with squared loss. Given a dataset $\{\data,\vec{y}\}$, the regularized training problem is as follows
\begin{align}\label{eq:newmodel_primal1}
        P^*=\min_{\theta \in \Theta} \frac{1}{2}\left \| f_{\theta}(\data) -\vec{y}\right\|_2^2+\frac{\beta}{2} \sum_{k=1}^{K}\left( \|\weight_{2k}\|_2^2+\weightscalar_{3k}^2\right),
\end{align}
where $\Theta:=\{\theta : \|\weightmat_{1k}\|_F\leq 1, \; \forall k \in [K]\}$ and
\begin{align*}
    f_{\theta}(\data):=\sum_{k=1}^{K}f_{\theta,k}(\data) .
\end{align*}
Using the rescaling in Lemma \ref{lemma:scaling_deep_overview}, \eqref{eq:newmodel_primal1} can be written as
\begin{align}\label{eq:newmodel_primal2}
           P^*=\min_{\theta \in \Theta_p} \ \frac{1}{2}\left \| f_{\theta}(\data) -\vec{y}\right\|_2^2+\beta \|\weight_3\|_1.
\end{align}
We then take the dual with respect to $\weight_3$ and then change the order of min-max to obtain the following dual problem
\begin{align}\label{eq:newmodel_dual1}
          P^* \geq &D^*:= \max_{\dual} -\frac{1}{2}\|\dual-\vec{y}\|_2^2+\frac{1}{2}\|\vec{y}\|_2^2 \\ \nonumber
          &\text{ s.t. } \max_{\theta \in \Theta_p}  \left|\dual^T \relu{\relu{\data \weightmat_{1}}\weight_{2}}\right|\leq \beta.
\end{align}
In order to obtain the bidual of \eqref{eq:newmodel_primal1}, we again take the dual of \eqref{eq:newmodel_dual1} with respect to $\dual$, which yields
\begin{align}\label{eq:newmodel_dual2_bidual}
 &   P_B^*:= \min_{\boldsymbol{\mu}}\frac{1}{2}\left\|\int_{ \theta \in \Theta_p } \relu{\relu{\data\weightmat_{1}}
    \weight_{2}}d{\mu}(\theta)-\vec{y}\right\|_2^2 \nonumber \\
&\hspace{5.5cm} +\beta \|\boldsymbol{\mu}\|_{TV}, 
\end{align}
where $\|\boldsymbol{\mu}\|_{TV}$ is the total variation norm of the Radon measure $\boldsymbol{\mu}$. We now note that \eqref{eq:newmodel_dual2_bidual} is an infinite-size regularized neural network training problem studied in \cite{bach2017breaking}, and it is convex. Therefore, strong duality holds, i.e., $D^*=P_B^*$. We also note that although \eqref{eq:newmodel_dual2_bidual} involves an infinite dimensional integral, by Caratheodory's theorem, this integral form can be represented as a finite summation of at most $n+1$ Dirac delta measures \cite{rosset2007}. Thus, we select ${\boldsymbol{\mu}}^\prime = \sum_{i=1}^{K^*} \weightscalar_{3i} \ \delta(\theta-\theta_i)$, where $K^* \leq n+1$, to achieve the following finite-size problem
\begin{align}\label{eq:newmodel_dual5_bidual}
     P_B^*=\min_{\theta \in \Theta_p}\frac{1}{2}\left\| \sum_{i=1}^{K^*}f_{\theta,i}(\data)-\vec{y}\right\|_2^2+\beta \|\weight_{3}\|_1.
\end{align}
Note that \eqref{eq:newmodel_dual5_bidual} is the same problem with \eqref{eq:newmodel_primal2} provided that $K \geq K^*$. Therefore, strong duality holds, i.e., $P^*=P_B^*=D^*$. Using strong duality, we first characterize optimal hidden layer weights via the active constraints of the dual problem. We then introduce a novel framework to represent the constraints in a convex form to obtain an equivalent convex formulation for the primal problem \eqref{eq:newmodel_primal1}.

We now represent the constraint in \eqref{eq:newmodel_dual1} as
\begin{align*}
&\left\{\dual: \max_{\theta \in \Theta_p}  \dual^T \relu{\relu{\data \weightmat_1}\weight_2}  \le \beta \right\}  \\
&\bigcap \left\{\dual:\max_{\theta \in \Theta_p}    -\dual^T \relu{\relu{\data \weightmat_1}\weight_2}\le \beta  \right\}.
\end{align*}
We first focus on a single-sided dual constraint
\begin{align}
\max_{\theta \in \Theta_p} \, \dual^T \relu{\relu{\data \weightmat_1}\weight_2} \le \beta \label{eq:newmodel_dualconst1}.
\end{align}
Noting that $\Theta_p=\{\theta \in \Theta: \|\weightmat_{1k}\|_F\leq 1,\;\|\weight_{2k}\|_2\leq 1, \forall k \in [K]\}$, we then equivalently write the constraint in \eqref{eq:newmodel_dualconst1} as
\begin{align}
    \max_{\mathcal{I}_j \in \{\pm1\}}\max_{\substack{\theta \in \Theta_p \\ \weight_2 \geq 0}} \, \dual^T \relu{\sum_{j=1}^{m_1}\mathcal{I}_j\relu{\data \weight_{1j}\weightscalar_{2j}}} \le \beta, \label{eq:newmodel_dualconst2}
\end{align}
where $\mathcal{I}_{j}=\text{sign}(\weightscalar_{2j})\in\{+1,-1\}$. Now, modifying $\|\weightmat_1\|_F\leq 1$ as $\|\weight_{1j}\|_2^2 \leq t_j$ such that $\vec{1}^T\vec{t}\leq 1$ and defining $\weight_{1j}^\prime=\sqrt{\weightscalar_{2j}^\prime}\weight_{1j}$, where $\weightscalar_{2j}^\prime=\weightscalar_{2j}^2$, yield
\begin{align}
   \max_{\substack{t_j\geq 0\\ \vec{1}^T\vec{t}\leq 1\\\mathcal{I}_j \in \{\pm1\}}}\max_{\substack{\|\weight_{1j}^{\prime}\|_2^2/w_{2j}^\prime \leq  t_j\\ \vec{1}^T\weight_2^\prime\leq 1\\ \weight_2^\prime \geq 0}} \, \dual^T \relu{\sum_{j=1}^{m_1}\mathcal{I}_j\relu{\data \weight_{1j}^\prime}} \le \beta. \label{eq:newmodel_dualconst3}
\end{align}
We remark that \eqref{eq:newmodel_dualconst3} is non-convex due to the ReLU activation. Therefore, to eliminate ReLU without altering the constraints, we introduce a notion of \emph{hyperplane arrangement} as follows.

Let $\mathcal{H}_1$ and $\mathcal{H}_2$ be the sets of all hyperplane arrangements for the hidden layers, which are defined as 
\begin{align*}
&\mathcal{H}_1 := \bigcup \big \{ \{\sign(\data \weight)\} : \weight \in \real^d \big \}\\
&\mathcal{H}_2 := \bigcup \big \{ \{\sign(\relu{\data \weightmat_1}\weight_2)\} : \weightmat_1 \in \mathbb{R}^{d\times m_1},\\ & \hspace{5.2cm}\weight_2 \in \mathbb{R}^{m_1} \big \}.
\end{align*}
%
%It is obvious that the sets $\mathcal{H}_1$ and $\mathcal{H}_2$ are bounded, i.e., $\exists N_H \in \mathbb{N} <\infty$ such that $\vert \mathcal{H}_1 \vert,\vert \mathcal{H}_2 \vert \le N_H$. 
We next define an alternative representation of the sign patterns in $\mathcal{H}_1$ and $\mathcal{H}_2$, which is the collection of sets that correspond to positive signs for each element in $\mathcal{H}_j$ as follows
\begin{align*}
&\mathcal{S}_{j} := \big\{ \{ \cup_{h_i=1} \{i\}  \} : \vec{h} \in \mathcal{H}_j \big\}, \forall j \in [2].
\end{align*}
We note that ReLU is an elementwise function that masks the negative entries of a vector/matrix. Hence, we define two diagonal mask matrices $\diag(S_j) \in \real^{n\times n}$ as $\diag(S_j)_{ii} :=  \mathbf{1} [ i \in S_j] $. We now enumerate all hyperplane arrangements and signs, and index them in an arbitrary order, which are denoted as $\mathcal{I}_j$, $\diag_{1ij}$, and $\diag_{2l}$, where $i \in [P_1]$, $l \in [P_2]$, $P_1=|\mathcal{S}_1|$, and $P_2=| \mathcal{S}_2|$. We then rewrite \eqref{eq:newmodel_dualconst3} as
\begin{align*}
&\max_{\substack{i \in [P_1]\\ l \in [P_2]}}\max_{\substack{t_j\geq 0\\ \vec{1}^T\vec{t}\leq 1\\\mathcal{I}_j \in \{\pm1\}}}\max_{\substack{\|\weight_{1j}^{\prime}\|_2^2/w_{2j}^\prime \leq  t_j\\ \vec{1}^T\weight_2^\prime\leq 1\\ \weight_2^\prime \geq 0}} \, \dual^T \diag_{2l} \sum_{j=1}^{m_1}\mathcal{I}_j\diag_{1ij}\data \weight_{1j}^\prime \\
    &\text{ s.t. } (2\diag_{1ij}) -\vec{I}_n)\data \weight_{1j}^\prime\geq 0,\,\forall i,j, \\& (2\diag_{2l} -\vec{I}_n)\sum_{j=1}^{m_1} \mathcal{I}_j\diag_{1ij}\data \weight_{1j}^\prime \geq 0,\,\forall i, l,  \nonumber
\end{align*}
where we use an alternative representation for ReLU as $\relu{\data\weight_1}=\diag \data \weight_1$ provided that $(2\diag-\vec{I}_n)\data \weight_1 \geq 0$. Therefore, we can convert the non-convex dual constraints in \eqref{eq:newmodel_dualconst1} to a convex constraint given fixed diagonal matrices $\{\diag_{1ij}\}_{j=1}^{m_1} $, $\diag_{2l}$ and a fixed set of signs $\{\mathcal{I}_j\}_{j=1}^{m_1}$ (see Section \ref{sec:proof_maintheo} for details). 

Using this new representation for the dual constraints, we then take the dual of \eqref{eq:newmodel_dual1} to obtain the convex bidual form of the primal problem \eqref{eq:newmodel_primal1} as described in the next theorem.

\begin{theos} \label{theo:main_theorem}
The non-convex training problem in \eqref{eq:newmodel_primal1} can be equivalently stated as a convex problem as follows
\begin{align}
\label{eq:newmodel_final}
  &\min_{\vec{w},\vec{w}^{\prime} \in \mathcal{C}} \frac{1}{2}\left\|\tilde{\data}  \left(\vec{w}^{\prime}-\vec{w}\right)-\vec{y}   \right\|_2^2+\beta  \left( \|\vec{w}\|_{2,1}+\|\vec{w}^{\prime}\|_{2,1}\right) 
\end{align}
where $\|\cdot\|_{2,1}$ is $d$ dimensional group norm operator such that given a vector $\vec{u} \in \mathbb{R}^{d P}$, $\|\vec{u}\|_{2,1}:= \sum_{i=1}^P \|\vec{u}_i\|_2$, where $\vec{u}_i$'s are the ordered $d$ dimensional partitions of $\vec{u}$. Moreover, $\tilde{\data} \in \mathbb{R}^{n \times 2dm_1P_1 P_2}$ and $\mathcal{C}$ are defined as
\begin{align*}
   & \mathcal{C}:=\Big\{\vec{w}\in \mathbb{R}^{2 d m_1 P_1 P_2} \;: \; \\
   &(2 \diag_{1ij}-\vec{I}_n) \data\vec{w}_{ijl}^{+}\geq 0, \;(2 \diag_{1ij}-\vec{I}_n) \data\vec{w}_{ijl}^{-}\leq 0,\, \\
  & (2 \diag_{2l}-\vec{I}_n)  \diag_{1ij} \data\vec{w}_{ijl}^{\pm}\geq 0, \forall i,j,l,\pm   \Big\} \\
  &\tilde{\data}:= \begin{bmatrix} \tilde{\data}_s & \vec{0}\\
  \vec{0} & \tilde{\data}_s
  \end{bmatrix},
\end{align*}
where $\vec{w}, \vec{w}^{\prime} \in \mathbb{R}^{2 d m_1 P_1 P_2}$ are the vectors constructed by concatenating $\{\{\{\{\vec{w}_{ijl}^{\pm}\}_{i=1}^{P_1}\}_{j=1}^{m_1}\}_{l=1}^{P_2}\}_{\pm}$ and $\{\{\{\{\vec{w}_{ijl}^{\pm^\prime}\}_{i=1}^{P_1}\}_{j=1}^{m_1}\}_{l=1}^{P_2}\}_{\pm}$, respectively, and
\begin{align*}
     \tilde{\data}_s:= \begin{bmatrix}\diag_{21}  \diag_{111}\data\, \ldots \, \diag_{2l}  \diag_{1ij}\data \,\ldots \,\diag_{2P_2}  \diag_{1P_1 m_1} \data\end{bmatrix}.
\end{align*}
\end{theos}

\vspace{1in}
\textbf{Difference between convex programs for two-layer \cite{pilanci2020convex} and three-layer (ours) networks:} \cite{pilanci2020convex} introduced a convex program for two-layer networks. In that case, since there is only a single ReLU layer, the data matrix $\data$ is multiplied with a single hyperplane arrangement matrix, namely $\diag_i$, and the effective high dimensional data matrix becomes $\tilde{\data}_s=[\diag_1 \data \,\ldots \,\diag_P \data]$. However, our architecture in \eqref{eq:newmodel_primal1} has two ReLU layers, combination of which can generate significantly more complex features, which are associated with local variables $\{\vec{w}_{ijl}^{\pm}\}$ that interact with data through the multiplication of two diagonal matrices as $\tilde{\data}_s= [\diag_{21}  \diag_{111}\data\, \ldots \, \diag_{2P_2}  \diag_{1P_1 m_1} \data]$.  In particular, a deep network can be precisely interpreted as a high-dimensional feature selection method due to convex group sparsity regularization, which encourages a parsimonious model. In simpler terms, such deep ReLU networks are group lasso models with additional linear constraints. Therefore, our result reveals the impact of having additional layers and its implications on the expressive power of a network.

\begin{figure}[h]
        \centering
            \centering
            \includegraphics[width=0.4\textwidth,height=0.4\textwidth]{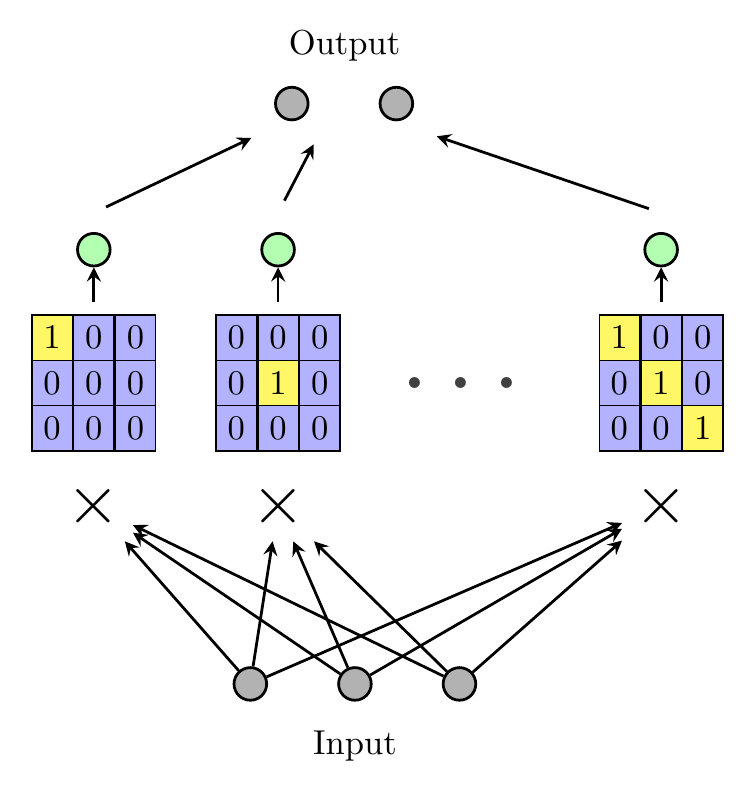}
            \caption{Equivalent convex model for the network in Figure \ref{fig:subnetworks}. Here, nonlinear ReLU layers are replaced with a linear layer and a transformation that maps to the input data to higher dimensional space by multiplying it with fixed diagonal matrices as detailed in Theorem \ref{theo:main_theorem}.  }\label{fig:convex_eq}%\vskip -0.1in
    \end{figure}

The next result shows that there is a direct mapping between the solutions to \eqref{eq:newmodel_primal1} and \eqref{eq:newmodel_final}. Therefore, once we solve the convex program in \eqref{eq:newmodel_final}, one can construct the optimal network parameters in their original form in \eqref{eq:newmodel_primal1} (see Section \ref{sec:proof_mapping} for the explicit definitions of the mapping).
\begin{props} \label{prop:weight_construction}
An optimal solution to the non-convex training problem in \eqref{eq:newmodel_primal1}, i.e., $\{\weightmat_{1k}^*,\weight_{2k}^*,\weightscalar_{3k}^*\}_{k=1}^K$, can be constructed using the optimal solution to \eqref{eq:newmodel_final}, denoted as $(\vec{w}^*,\vec{w}^{\prime^*})$. Therefore, there is a direct mapping between the architectures in Figure \ref{fig:subnetworks} and \ref{fig:convex_eq}.
\end{props}

%%%%%%% remark %%%%%%
\begin{rems}\label{rem:arrangment_samp}
The problem in \eqref{eq:newmodel_final} can be approximated by sampling a set of diagonal matrices $\{\{\diag_{1ij}\}_{i=1}^{\bar{P}_1}\}_{j=1}^{m_1}$ and $\{\diag_{2l}\}_{l=1}^{\bar{P}_2}$. As an example, one can generate random vectors $\vec{u}_{ij}$'s from an arbitrary distribution, e.g., $\vec{u}_{ij}\sim \mathcal{N}(\vec{0},\vec{I}_d)$, $\bar{P}_1$ times and then let $\diag_{1ij}=\text{diag}(\mathbf{1}[\data \vec{u}_{ij}\geq 0]), \forall i \in [\bar{P}_1]$, $\forall j \in [m_1]$.  Similarly, one can randomly generate $\vec{U}_{1} \in \mathbb{R}^{d \times m_1}$ and $\vec{u}_2 \in \mathbb{R}^{ m_1}$ $\bar{P}_2$ times and then let $\diag_{2l}=\text{diag}\left(\mathbf{1}\left[\relu{ \data \vec{U}_{1l}}\vec{u}_{2l}\geq 0\right] \right), \forall l \in [\bar{P}_2]$. Then, one can solve the convex problem in \eqref{eq:newmodel_final} using these hyperplane arrangements. In fact, SGD applied to the non-convex problem in \eqref{eq:newmodel_primal1} can be viewed as an active set optimization strategy to solve the equivalent convex problem, which maintains a small active support. We also note that global optimums of the convex problem are the fixed points for SGD, i.e., stationary points of \eqref{eq:newmodel_primal1}. Furthermore, one can bound the suboptimality of any solution found by SGD for the non-convex problem using the dual of \eqref{eq:newmodel_final}.
\end{rems}

%%%%%%%% synthetic dataset figure %%%%%%%%%%%
\begin{figure*}[ht]
        \centering
        \begin{subfigure}[b]{0.33\textwidth}
            \centering
            \includegraphics[width=\textwidth]{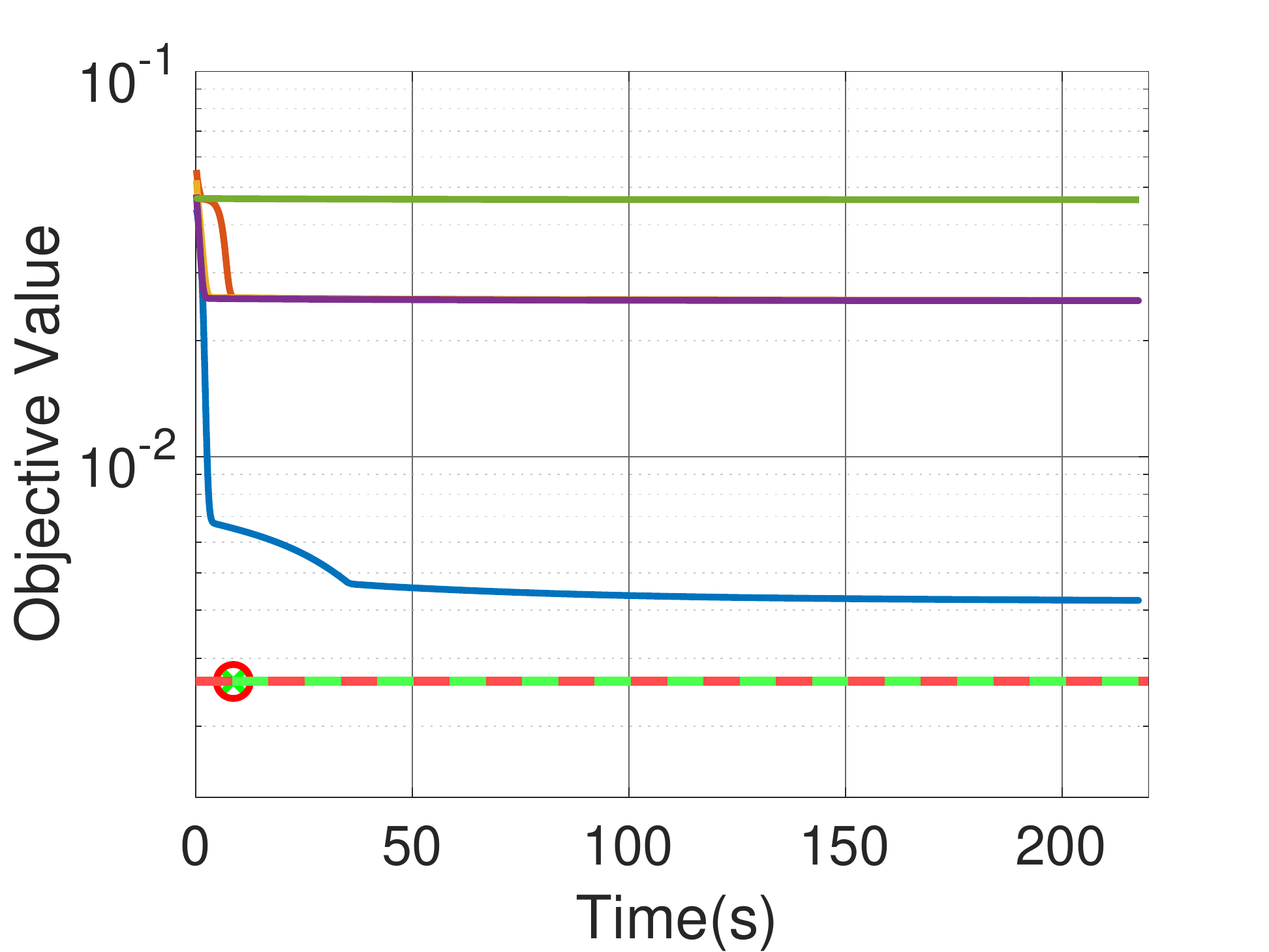}
            \caption{$K=2$}
        \end{subfigure}
        \hfill
        \begin{subfigure}[b]{0.33\textwidth}  
            \centering 
            \includegraphics[width=\textwidth]{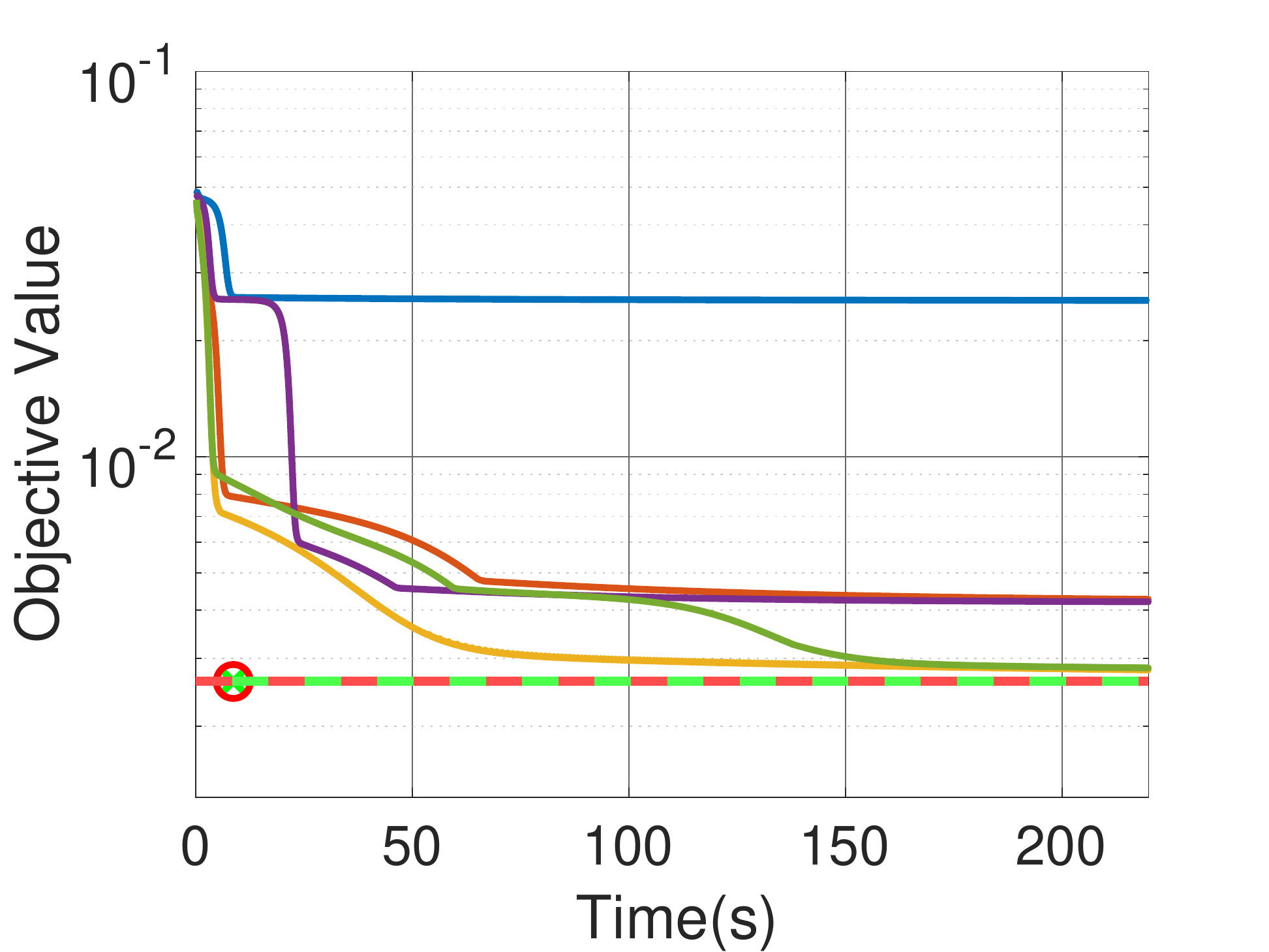}
            \caption{$K=5$}
        \end{subfigure}        \hfill
        \begin{subfigure}[b]{0.33\textwidth}   
            \centering 
            \includegraphics[width=\textwidth]{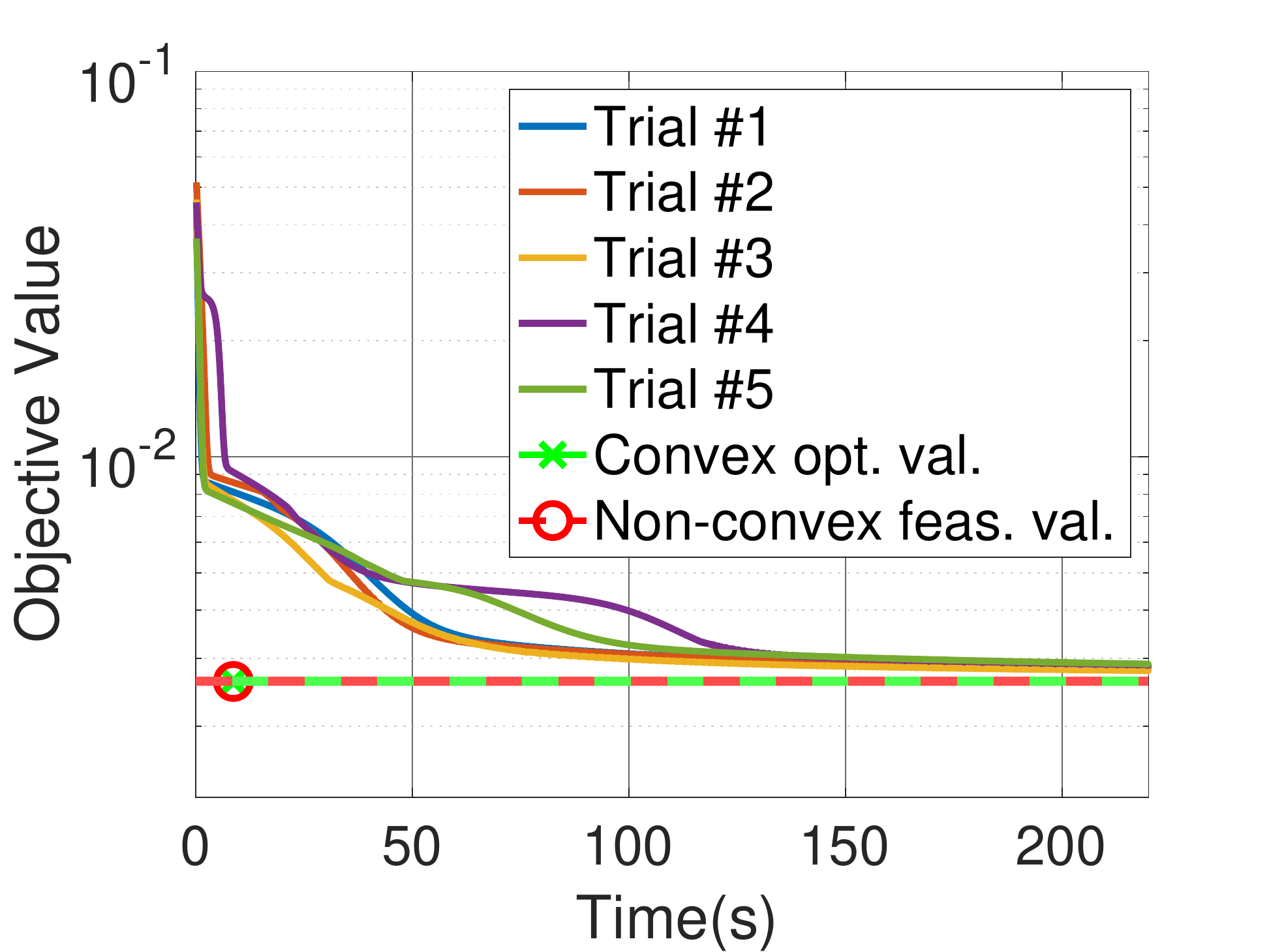}
            \caption{$K=15$}
        \end{subfigure}
	\caption{Training cost of a three-layer architecture trained with SGD (5 initialization trials) on a synthetic dataset with $(n,d,m_1,\beta,\text{batch size})=(5,2,3,0.002,5)$ , where the green line with a marker represents the objective value obtained by the proposed convex program in \eqref{eq:newmodel_final} and the red line with a marker represents the non-convex objective value in \eqref{eq:newmodel_primal1} of a classical ReLU network constructed from the solution of convex program as described in Proposition \ref{prop:weight_construction}. Here, we use markers to denote the total computation time of the convex optimization solver.}\label{fig:synthetic}
    %\vskip -0.2in
    \end{figure*}
\begin{table*}
\begin{center}
\resizebox{1.7\columnwidth}{!}{%
\begin{tabular}{|c|c|c|c|c|c|c|}
\hline
& \multicolumn{2}{|c|}{{\bf Dataset Size}} & \multicolumn{2}{|c|}{ {\bf Training Objective}} & \multicolumn{2}{|c|}{{\bf Test Error}} \\ \hline
  % \cmidrule(r){2-5}
 & $n$ & $d$  & \texttt{SGD} & \texttt{Convex} & \texttt{SGD} & \texttt{Convex}\\ \hline \hline
acute-inflamation \cite{acute_dataset} & 120 & 6 &  0.0029  & {\bf 0.0013}   & 0.0224 & {\bf 0.0217}\\ \hline
acute-nephritis \cite{acute_dataset}  & 120 & 6 &  0.0039 & {\bf 0.0021}  & 0.0198 & {\bf 0.0192}   \\ \hline
balloons \cite{uci_repository} & 16 & 4 & 0.7901 & {\bf 0.6695} & 0.2693 & {\bf 0.1496}  \\ \hline
breast-tissue \cite{ uci_repository} &106 & 9 & 0.5219 & {\bf 0.3979}  & 1.4082 & {\bf 1.0377}  \\ \hline
fertiliy \cite{uci_repository} & 100 & 9 & 0.125 & {\bf 0.1224}  & {\bf 0.3551} &  0.5050  \\ \hline
pittsburg-bridges-span \cite{uci_repository} & 92 & 7 & 0.1723 & {\bf 0.1668} & 1.4373 & {\bf 1.3112} \\ \hline
\end{tabular}%
}
\end{center}
\caption{Training objective and test error of a three-layer architecture trained using SGD and the convex program in \eqref{eq:newmodel_final}, i.e., denoted as ``Convex", on some small scale UCI datasets with $(m_1,K,\beta, \text{batch size})=(7,10,0.1,\lfloor n/8\rfloor)$.}\label{tab:uci_smallscale}%\vskip -0.2in
\end{table*}

%%%%%%%%%%%%%%%%%%%%%%%%%%%%%%% new subsection %%%%%%%%%%%%%%%%%%%%%%%%%%%%%%%%%%%%%%%%%
\subsection{Multilayer Hyperplane Arrangements  }
It is known that the number of hyperplane arrangements for the first layer, i.e., $P_1$, can be upper-bounded as follows
\begin{align}\label{eq:threelayer_hyperplane1}
    P_1\le 2 \sum_{k=0}^{r-1} {n-1 \choose k}\le 2r\left(\frac{e(n-1)}{r}\right)^r =\mathcal{O}(n^{ r})\,
\end{align}
for $r\le n$, where $r:=\mbox{rank}(\data)\leq d$ \cite{ojha2000enumeration,stanley2004introduction,winder1966partitions,cover1965geometrical}. 
 For the second layer, we first note that the activations before ReLU can be written as follows
\begin{align*}
\sum_{j=1}^{m_1} \relu{\data \weight_{1j}}\weightscalar_{2j}=\sum_{j=1}^{m_1} \relu{\data \weight_{j}^\prime}\mathcal{I}_{j}  =\sum_{j=1}^{m_1}\mathcal{I}_{j} \diag_{1j}\data \weight_{j}^\prime,
\end{align*}
which can also be formulated as a matrix-vector product 
\begin{align*}
   & \underbrace{\begin{bmatrix} \mathcal{I}_{1}\diag_{11} \data &
    \mathcal{I}_{2}\diag_{12} \data&
    \ldots&
    \mathcal{I}_{m_1}\diag_{1m_1} \data
    \end{bmatrix} }_{\data^\prime} \underbrace{\mathrm{vec}(\{\weight_j^\prime\}_{j=1}^{m_1}\})}_{\weight^\prime} \\&=\data^\prime \weight^\prime,
\end{align*}
where $\data^\prime \in \mathbb{R}^{n \times m_1d}$ and $\weight^\prime \in \mathbb{R}^{m_1 d}$. Therefore, given a fixed set $\{\mathcal{I}_{j},\diag_{1j}\}_{j=1}^{m_1}$, the number of hyperplane arrangements for $\data^\prime$ can be upper-bounded as follows
\begin{align*}
    P_2^\prime \leq  2r^\prime\left(\frac{e(n-1)}{r^\prime}\right)^{r^\prime}\leq 2m_1 r\left( \frac{e(n-1)}{m_1 r}\right)^{m_1 r},
\end{align*}
where $r^\prime:=\mathrm{rank}(\data^\prime)\leq m_1 r $ since $\mathrm{rank}(\data)=r$ and  we assume that $m_1r\leq n$. Since there exist 2 and $P_1$ possible choices for each $\mathcal{I}_{j}$ and $ \diag_{1j}$, respectively, the total number of hyperplane arrangements for the second layer can be upper-bounded as follows
\begin{align}\label{eq:threelayer_hyperplane2}
P_2 \leq P_2^\prime (2 P_1)^{m_1} &\leq m_1 r 2^{m_1+1} P_1^{m_1}  \left( \frac{e(n-1)}{m_1 r}\right)^{m_1 r}  \nonumber \\
&\leq \frac{2^{2m_1+1}(e(n-1))^{2m_1r}}{m_1^{m_1 r -1} r^{2m_1r-m_1-1}} \nonumber\\
&=\mathcal{O}(n^{m_1 r}),
\end{align}
which is polynomial in $n$ and $d$ since $m_1$ and $r$ are fixed scalars.

\begin{rems}
For convolutional networks, we operate on the patch matrices $\{\data_b\}_{b=1}^B$ instead of $\data$, where $\data_b \in \mathbb{R}^{n \times h}$ and $h$ denotes the filter size. Therefore, even when the data matrix is full rank, i.e., $r=d$, the number of relevant hyperplane arrangements in \eqref{eq:threelayer_hyperplane1} is $\mathcal{O}(n^{ r_c})$, where $r_c:= \mbox{rank}([\data_1; \ldots; \data_B])\leq h \ll d$. As an example, if we consider a convolutional network with $m_1$ $3 \times 3$ filters, then $r_c \leq 9$ independent of the dimension $d$. As a corollary, this shows that the parameter sharing structure in CNNs significantly limits the number of hyperplane arrangements after ReLU activation, which might be one of the key factors behind their generalization performance in practice.
\end{rems}
%%%%%%%%% remark %%%%%%%%
\begin{rems} \label{rem:hyperplane_arrangments}
We can also compute the number of the hyperplane arrangements in the $l^{th}$ layer, i.e., $P_{l}$. We first note that if we use the same approach for $P_3$ then due to the multiplicative structure in \eqref{eq:threelayer_hyperplane2}, we have $P_3 \leq P_3^\prime (2P_2)^{m_2} \leq \mathcal{O}(n^{m_2m_1  r})$. Therefore, applying this relation recursively yields $P_{l}\leq \mathcal{O}(n^{r \prod_{j=1}^{l-1}m_j})$, which is also a polynomial term in both $n$ and $d$ for fixed data rank $r$ and fixed width $\{m_j\}_{j=1}^{l-1}$.
\end{rems}

%%%%%%%% cifar-mnist figure %%%%%%%%%%%
\begin{figure*}[t]
        \centering
        \begin{subfigure}[b]{0.45\textwidth}  
            \centering 
         \includegraphics[width=.9\textwidth,height=.7\textwidth]{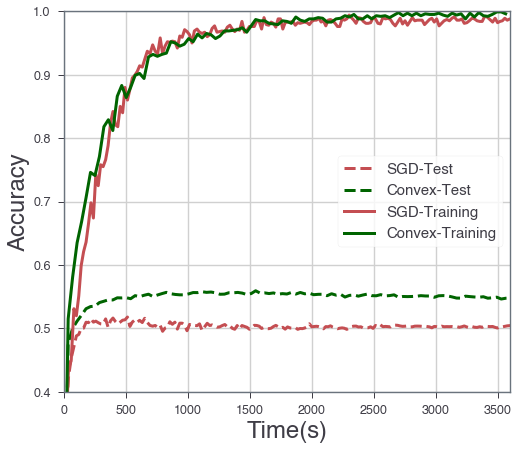}
            \caption{CIFAR-10}\label{fig:cifar}
        \end{subfigure}        \hfill
        \begin{subfigure}[b]{0.45\textwidth}   
            \centering 
          \includegraphics[width=.9\textwidth,height=.7\textwidth]{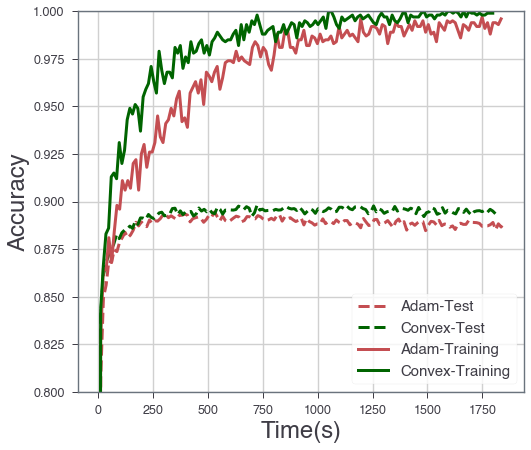}
            \caption{Fashion-MNIST}\label{fig:fmnist}
        \end{subfigure}
	\caption{Accuracy of a three-layer architecture trained using the non-convex formulation \eqref{eq:newmodel_primal1} and the proposed convex program \eqref{eq:newmodel_final}, where we use (a) CIFAR-10 with $(n,d,m_1,K,\beta,\text{batch size})=(5\text{x} 10^4,3072,100,40,10^{-3},10^{3})$ and (b) Fashion-MNIST with $(n,d,m_1,K,\beta,\text{batch size})=(6\text{x} 10^4,784,100,40,10^{-3},10^3)$. We note that the convex model is trained using (a) SGD and (b) Adam and we use the approximation in Remark \ref{rem:arrangment_samp} for the convex programs. }\label{fig:cifar_mnist}
    %\vskip -0.1in
    \end{figure*}

%%%%%%%%%%%%%%%%%%%%%%%%%%%%%%% new subsection %%%%%%%%%%%%%%%%%%%%%%%%%%%%%%%%%%%%%%%%%
\subsection{Training Complexity  }
In this section, we analyze the computational complexity to solve the convex problem in \eqref{eq:newmodel_final}. We first note that \eqref{eq:newmodel_final} has $4dm_1 P_1 P_2$ variables and $4nP_1P_2(m_1+1)$ constraints. Thus, given the bound on the hyperplane arrangements in \eqref{eq:threelayer_hyperplane1} and \eqref{eq:threelayer_hyperplane2}, a standard convex optimization solver, e.g., an interior point method, can globally optimize \eqref{eq:newmodel_final} with a polynomial-time complexity, i.e., $\mathcal{O}(d^3 m_1^3 n^{3(m_1+1)r})$. This result might also be extended to arbitrarily deep networks as detailed below.

\begin{cors}
Remark \ref{rem:hyperplane_arrangments} shows that $L$-layer architectures can be globally optimized
with $\mathcal{O}\left(d^3 \left(\prod_{j=1}^{L-2}m_j^3\right) n^{3r \left(1+\sum_{l=1}^{L-2}\prod_{j=1}^{l}m_j\right)} \right)$ complexity, which is polynomial in $n,d$ for fixed rank and widths.
\end{cors}

%%%%%%%%%%%%%%%%%%%%%%%%%%%%%%% new section %%%%%%%%%%%%%%%%%%%%%%%%%%%%%%%%%%%%%%%%%
\section{Numerical Results}

In this section\footnote{Additional experiments and details on the numerical results can be found in Section \ref{sec:numerical_supp} the supplementary file.}, we present several numerical experiments validating our theory in the previous section. We first conduct an experiment on a synthetic dataset with $(n,d)=(5,2)$. For this dataset, we first randomly generate $d$ dimensional data samples $\{\sample_i\}_{i=1}^n$ using a multivariate Gaussian distribution with zero mean and identity covariance, i.e., $\sample_i \sim \mathcal{N}(\vec{0},\vec{I}_d)$. We then forward propagate these samples through a randomly initialized three-layer architecture with $m_1=3$ and $K=5$ to obtain the corresponding labels $\vec{y} \in \mathbb{R}^n$. We then train the three-layer architecture in \eqref{eq:newmodel_primal1} on this synthetic dataset using SGD and our convex approach in \eqref{eq:newmodel_final}. In Figure \ref{fig:synthetic}, we plot the training objective values with respect to the computation time taken by each algorithm, where we include $5$ independent initialization trials for SGD. Moreover, for the convex approach, we plot both the objective value of the convex program in \eqref{eq:newmodel_final} and its non-convex equivalent constructed as described in Proposition \ref{prop:weight_construction}. Here, we observe that when $K$ is small, SGD trials tend to get stuck at a local minimum. Furthermore, as we increase the number of sub-networks, all the trials are able to converge to the global minimum achieved by the convex program. We also note that these observations are also consistent with the landscape visualizations in Figure \ref{fig:landscape_synthetic}.

% \begin{table*}
% \begin{center}
% \begin{tabular}{|c|c|c|c|c|c|c|c|c|c|c|}
% \hline
% & \multicolumn{5}{|c|}{Training Objective} & \multicolumn{5}{|c|}{Test Error} &\\ \hline
%   % \cmidrule(r){2-5}
%   & SGD & Convex & SGD & Convex\\ \hline
% SGD &  &  & &  &  &  & & & & \\ \hline
% Convex &  &  & &  &  &  & & & &    \\ \hline

% \end{tabular}
% \end{center}
% \caption{Results.   Ours is better.}
% \end{table*}

 In order to validate our theory, we also perform several experiments on some small scale real datasets available in UCI Machine Learning Repository \cite{uci_repository}. For these datasets, we consider a regression framework with $(m_1,K)=(7,10)$ and compare the training and test performance of SGD and the convex program in \eqref{eq:newmodel_final}. As reported in Table \ref{tab:uci_smallscale}, our convex approach achieves a lower training objective for all the datasets. Although the convex approach also obtains lower test errors for five out of six datasets, there is a case where SGD achieves better test performance, i.e., the fertility dataset in Table \ref{tab:uci_smallscale}. We believe that this is an interesting observation related to the generalization properties of SGD and the convex approach and leave this as an open problem for future research. 

We also conduct an experiment on CIFAR-10 \cite{cifar10} and Fashion-MNIST \cite{fashionmnist}. Here, we consider a ten class classification problem using an architecture with $(m_1,K)=(100,40)$, and report the test and training accuracies. In Figure \ref{fig:cifar_mnist}, we provide these values with respect to time. This experiment verifies the performance boost provided by training on the convex formulations.

%%%%%%%%%%%%%%%%%%%%%%%%%%%%%%% new section %%%%%%%%%%%%%%%%%%%%%%%%%%%%%%%%%%%%%%%%%
\section{Concluding Remarks}
We presented a convex analytic framework to characterize the optimal solutions to an architecture constructed by combining multiple deep ReLU networks in parallel. Particularly, we first derived an exact equivalent formulation for the non-convex primal problem using convex duality. This formulation has two significant advantages over the non-convex primal problem. First, since the equivalent problem is convex, it can be globally optimized by standard convex solvers without requiring any exhaustive search to tune hyperparameters, e.g., learning rate and initialization, or heuristics such as dropout. Second, we proved that globally optimizing the equivalent problem has polynomial-time complexity with respect to the number of samples $n$ and the feature dimension $d$. Therefore, we proved the polynomial-time trainability of regularized ReLU networks with more than two layers, which was previously known only for basic two-layer ReLU networks \cite{pilanci2020convex}. More importantly, since the equivalent problem is convex, one can achieve further interpretations and develop faster solvers by utilizing the tools in convex optimization. 

Our approach also revealed an implicit regularization structure behind the original non-convex training problem. This structure is known as group $\ell_1$-regularization that encourages sparsity between certain groups of parameters. As a corollary, the regularization in the convex problem implies that the original non-convex training problem achieves sparse solutions at global minima, where sparsity is over the number sub-networks. Our analysis also demystified mechanisms behind empirical observations regarding the convergence of SGD (see Figure \ref{fig:synthetic}) and loss landscape (see Figure \ref{fig:landscape_synthetic} and \cite{zhang2019multibranch,haeffele2017global}).

\textbf{Limitations:} We conclude with the limitations of this work and some open research problems:
\begin{itemize}
    \item The parallel architectures studied in this work has three layers (two ReLU layers). Notice that even though we already provided some complexity results for deeper architectures, deriving the corresponding convex representations still remain an open problem.
    \item Each sub-network of our parallel architecture in \eqref{eq:sub-network} restricts the last hidden layer weights to be vectors (i.e. $\weight_{(L-1)k} \in \mathbb{R}^{m_{L-2}}$) and output layers to be scalars (i.e. $\weightscalar_{Lk} \in \mathbb{R}$). Therefore, when $L=3$, the parallel, $\sum_{k}\relu{\relu{\data \weightmat_{1k}}\weight_{2k} }\weightscalar_{3k}$, is closer to two-layer standard networks than three-layer networks (except having two ReLU layers) in terms of expressive power. However, we believe that our convex approach can be extended to truly three-layer or deeper networks by slightly tweaking either the architecture $f_{\theta}(\data)$ or the regularization function $\mathcal{R}(\theta)$.
    \item In order to utilize convex duality, we put unit $\ell_2$-norm constraints on the first $L-2$ layer weights, which do not reflect the common practice. Therefore, we conjecture that weight decay regularization might not be the proper way of regularizing deep ReLU networks.
    \item When the data matrix is full rank, our approach has exponential-time complexity, which is unavoidable unless $\mathrm{P}=\mathrm{NP}$ as detailed in \cite{pilanci2020convex}.
    \item Finally, our convex approach can also be applied to various neural network architectures, e.g., CNNs \cite{ergen2020cnn}, batch normalization \cite{ergen2021bn}, generative adversarial networks (GANs) \cite{sahiner2021gan}, and autoregressive models \cite{vikul2021generative}, with some technical modifications, which we left for future work.
\end{itemize}
\section*{Acknowledgements}
This work was partially supported by the National Science Foundation under grants IIS-1838179 and ECCS-2037304, Facebook Research, Adobe Research and Stanford SystemX Alliance.

\bibliographystyle{icml2021}
\bibliography{references}

%%%%%%%%%%%%%%%%%%%%%%%%%%%%%%% Supplementary %%%%%%%%%%%%%%%%%%%%%%%%%%%%%%%%%%%%%%%%%

\clearpage
%\newpage
\onecolumn
\appendix
\addcontentsline{toc}{section}{Supplementary Material} % Add the appendix text to the document TOC
\part{Supplementary Material} % Start the appendix part
\parttoc % Insert the appendix TOC
\section{Supplementary Material}

%%%%%%%%%%%%%%%%%%%%%%%%%%%%%%% new subsection %%%%%%%%%%%%%%%%%%%%%%%%%%%%%%%%%%%%%%%%%
\subsection{Additional numerical results} \label{sec:numerical_supp}
In this section, we provide detailed information about our experiments.

We first note that for small scale experiments, i.e., Figure \ref{fig:synthetic} and Table \ref{tab:uci_smallscale}, we use CVX \cite{cvx} and CVXPY \cite{cvxpy,cvxpy_rewriting} with the SDPT3 solver \cite{tutuncu2001sdpt3} to solve convex optimization problems in \eqref{eq:newmodel_final}. Moreover, the training is performed on the CPU of a laptop with i7 processor and 16GB of RAM. For UCI experiments, we use the $80\%-20\%$ splitting ratio for the training and test sets. Moreover, the learning rate of SGD is tuned via a grid-search on the training split. Specifically, we try different values and choose the best performing learning rate on the validation datasets, which turns out to be $0.3$.

For larger scale experiment in Figure \ref{fig:cifar_mnist}, we use a GPU with 50GB of memory. In order to train the constrained convex program in \eqref{eq:newmodel_final}, we now introduce an unconstrained version of the convex program as follows
\begin{align}
    \label{eq:newmodel_final_uncons}
  &\min_{\vec{w},\vec{w}^{\prime} \in \mathbb{R}^{2 d m_1 P_1 P_2}} \frac{1}{2}\left\|\tilde{\data}  \left(\vec{w}^{\prime}-\vec{w}\right)-\vec{y}   \right\|_2^2+\beta  \left( \|\vec{w}\|_{2,1}+\|\vec{w}^{\prime}\|_{2,1}\right) + \rho \left(g_{\mathcal{C}}(\vec{w})+g_{\mathcal{C}}(\vec{w}^\prime)\right)
\end{align}
where $\rho >0$ is a trade-off parameter and 
\begin{align*}
    g_{\mathcal{C}}(\vec{w}):=\vec{1}^T  \sum_{i,j,l} \left(\relu{-(2 \diag_{1ij}-\vec{I}_n) \data\vec{w}_{ijl}^{+}}+\relu{2 \diag_{1ij}-\vec{I}_n) \data\vec{w}_{ijl}^{-}}\right)
    +\vec{1}^T \sum_{i,j,l,\pm}  \relu{-(2 \diag_{2l}-\vec{I}_n)  \diag_{1ij} \data\vec{w}_{ijl}^{\pm}} .
\end{align*}
Since the problem in \eqref{eq:newmodel_final_uncons} is in an unconstrained form, we can directly optimize its parameters using conventional local search algorithms such as SGD and Adam. Hence, we are able to use PyTorch to optimize the non-convex objective in \eqref{eq:newmodel_primal1} and the convex objective in \eqref{eq:newmodel_final_uncons} on the conventional benchmark datasets such as CIFAR-10 and Fashion-MNIST datasets with their original training and test splits. For the learning rates of SGD and Adam optimizer (applied to the non-convex formulations), we again follow the same grid-search technique and select $1$ and $0.01$ as the learning rates, respectively. For SGD, we also use momentum with a parameter of $0.9$. Moreover, for the convex programs in both cases, we select the number of hyperplane arrangements for the first and second layer such that $P_1 P_2=K$ and set the learning rate and the trade-off parameter as $10^{-6}$ and $\rho=0.01$, respectively. Then, we run the algorithms on the non-convex and convex formulations for $200,100$ epochs using SGD in Figure \ref{fig:cifar}. Similarly, we run the algorithms on the non-convex and convex formulations for $150,150$ epochs using Adam in Figure \ref{fig:fmnist}. We also note that for all of these experiments, we use an approximated form of the convex program detailed in Remark \ref{rem:arrangment_samp}. Therefore, we conjecture that one can even further improve the performance by either sampling more hyperplane arrangements or developing a technique to characterize the set of hyperplane arrangements that generalize well.

\begin{figure*}[t]
        \centering
        \begin{subfigure}[b]{0.45\textwidth}  
            \centering 
         \includegraphics[width=.9\textwidth,height=.7\textwidth]{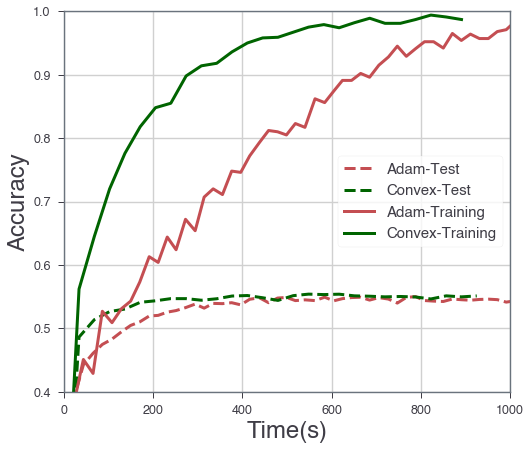}
            \caption{CIFAR-10}\label{fig:cifar_supp}
        \end{subfigure}        \hfill
        \begin{subfigure}[b]{0.45\textwidth}   
            \centering 
          \includegraphics[width=.9\textwidth,height=.7\textwidth]{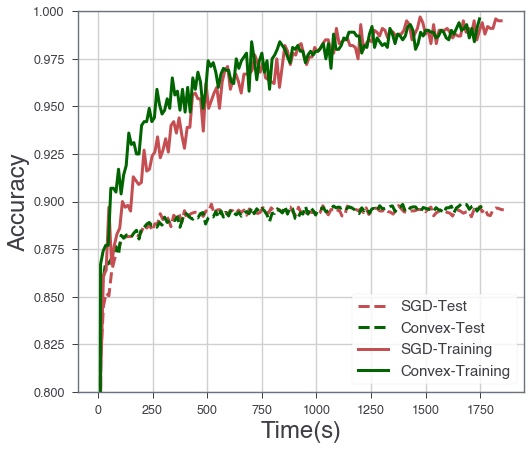}
            \caption{Fashion-MNIST}\label{fig:fmnist_supp}
        \end{subfigure}
	\caption{Accuracy of a three-layer architecture trained using the non-convex formulation \eqref{eq:newmodel_primal1} and the proposed convex program \eqref{eq:newmodel_final}, where we use (a) CIFAR-10 with $(n,d,m_1,K,\beta,\text{batch size})=(5\text{x} 10^4,3072,100,40,10^{-3},10^{3})$ and (b) Fashion-MNIST with $(n,d,m_1,K,\beta,\text{batch size})=(6\text{x} 10^4,784,100,40,10^{-3},10^3)$. We note that the convex model is trained using (a) Adam and (b) SGD. }\label{fig:cifar_mnist_supp}
    %\vskip -0.1in
    \end{figure*}

To complement the experiments in Figure \ref{fig:cifar_mnist}, we also conduct a new experiment, where we use Adam and SGD for CIFAR-10 and Fashion-MNIST, respectively. For this case, we use the same setup above except that the learning rates are chosen as $(5\text{x}10^{-7},5\text{x}10^{-4})$ and $(10^{-5},3)$ for CIFAR-10 and Fashion-MNIST respectively, where the former learning rates belong to the convex problems. We also run the algorithms on the non-convex and convex formulations for $66,26$ epochs using Adam in Figure \ref{fig:cifar_supp} and $150,150$ epochs using SGD in Figure \ref{fig:fmnist_supp}. We plot the accuracy values in Figure \ref{fig:cifar_mnist_supp}, where the training on the convex formulation achieves faster convergence and higher (or at least the same) accuracies compared to the training on the original non-convex formulation.

%%%%%%%% cifar-mnist figure %%%%%%%%%%%

%%%%%%%%%%%%%%%%%%%%%%%%%%%%%%% new subsection %%%%%%%%%%%%%%%%%%%%%%%%%%%%%%%%%%%%%%%%%
\subsection{Proof of Lemma \ref{lemma:scaling_deep_overview} }\label{sec:proof_scaling}
We first note that similar proofs are also presented in \cite{neyshabur_reg,infinite_width,ergen2019cutting,ergen2020aistats,ergen2020journal,ergen2020workshop,ergen2021revealing}.

For any $\theta \in \Theta$, we can rescale the parameters as $\bar{\weight}_{(L-1)k}=\alpha_k\weight_{(L-1)k}$ and $\bar{\weightscalar}_{Lk}= \weightscalar_{Lk}/\alpha_k$, for any $\alpha_k>0$. Then, the network output becomes
\begin{align*}
    f_{\bar{\theta},k}(\data)= \relu{\relu{\data \weightmat_{1k}} \ldots \bar{\weight}_{(L-1)k} }\bar{\weightscalar}_{Lk}= \relu{\relu{\data \weightmat_{1k}} \ldots \weight_{(L-1)k} }\weightscalar_{Lk},
\end{align*}
which proves $f_{\theta,k}(\data)=f_{\bar{\theta},k}(\data)$, $\forall k \in [K]$. In addition to this, we have the following basic inequality
\begin{align*}
   \frac{1}{2} \sum_{k=1}^K (\weightscalar_{Lk}^2+\| \weight_{(L-1)k}\|_2^2) \geq \sum_{k=1}^K (|\weightscalar_{Lk}| \text{ }\| \weight_{(L-1)k}\|_2),
\end{align*}
where the equality is achieved with the scaling choice $\alpha_k=\big(\frac{|\weightscalar_{Lk}|}{\| \weight_{(L-1)k}\|_2}\big)^{\frac{1}{2}}$ is used. Since the scaling operation does not change the right-hand side of the inequality, we can set $\|\weight_{(L-1)k} \|_2=1, \forall k \in [K]$. Therefore, the right-hand side becomes $\| \weight_L\|_1$.

Now, let us consider a modified version of the problem, where the unit norm equality constraint is relaxed as $\| \weight_{(L-1)k} \|_2 \leq 1$. Let us also assume that for a certain index $k$, we obtain  $\| \weight_{(L-1)k} \|_2 < 1$ with $\weightscalar_{Lk}\neq 0$ as an optimal solution. This shows that the unit norm inequality constraint is not active for $\weight_{(L-1)k}$, and hence removing the constraint for $\weight_{(L-1)k}$ will not change the optimal solution. However, when we remove the constraint, $\| \weight_{(L-1)k}\|_2 \rightarrow \infty$ reduces the objective value since it yields $\weightscalar_{Lk}=0$. Therefore, we have a contradiction, which proves that all the constraints that correspond to a nonzero $\weightscalar_{Lk}$ must be active for an optimal solution. This also shows that replacing $\|\weight_{(L-1)k}\|_2=1$ with $\| \weight_{(L-1)k} \|_2 \leq 1$ does not change the solution to the problem.

%%%%%%%%%%%%%%%%%%%%%%%%%%%%%%% new subsection %%%%%%%%%%%%%%%%%%%%%%%%%%%%%%%%%%%%%%%%%
\subsection{Constraint on the layer weights (Remark \ref{remark:regularization}) }\label{sec:proof_remark2}
Here, we prove that changing the unit norm constraint on the first $L-2$ layer weights does not change the structure of the regularization induced by the primal problem \eqref{eq:problem_def_overview}.

We first note that due to the AM-GM inequality, for each sub-network $k$, we have
\begin{align*}
    \sum_{l=1}^L \|\weightmat_{lk}\|_F^2 \geq L \prod_{l=1}^L \|\weightmat_{lk}\|_F^{2/L}
\end{align*}
where the equality is achieved when all the layer weight have the same Frobenius norm, i.e., $\|\weightmat_{1k}\|_F=\ldots=\|\weightmat_{Lk}\|_F$. Therefore, for a given set of arbitrary weight matrices, one can scale them such that their Frobenius norms are equal to each other and reduce the objective function in \eqref{eq:supp_problem_def2}. Based on this observation, for the rest of the derivations, we assume $\|\weightmat_{1k}\|_F=\ldots =\|\weightmat_{(L-2)k}\|_F=s$ without loss of generality. 

Now, let us consider the following primal problem instead of \eqref{eq:problem_def_overview}
\begin{align}
    P^*=& \min_{\theta \in \Theta} \mathcal{L}\left( \sum_{k=1}^K f_{\theta,k}(\data),\vec{y}\right)+ \frac{\beta}{2} \sum_{k=1}^K\sum_{l=L-1}^L \|\weightmat_{lk}\|_F^2 , \label{eq:supp_problem_def2}
\end{align}
where $\Theta:=\{\theta: \|\weightmat_{lk}\|_F \leq s, \forall l \in [L-2], \forall k \in [K]\}$. For this problem, we can follow all the derivations in the proof of Theorem \ref{theo:main_theorem} (see Appendix \ref{sec:supp_convexprogram} below) by changing $\vec{1}^T \vec{t}=1$ as $\vec{1}^T \vec{t}=s^2$ in \eqref{eq:newmodel_dualconst4}. Then, we will have an additional $s$ factor in the last step of \eqref{eq:newmodel_dualconst5}. This change will yield $\beta/s$ instead $\beta$ in \eqref{eq:newmodel_dualconst5}. Therefore, if we define a new variable as $\beta^\prime=\beta/s$, following the remaining steps in the proof below will yield the same convex program in \eqref{eq:newmodel_final_proof} with the regularization parameter $\beta^\prime$. Hence, the impact of the norm constraint in the primal problem \eqref{eq:supp_problem_def2} can be reverted by simply setting a new regularization parameter $\beta^\prime=\beta/s$.

%%%%%%%%%%%%%%%%%%%%%%%%%%%%%%% new subsection %%%%%%%%%%%%%%%%%%%%%%%%%%%%%%%%%%%%%%%%%
\subsection{Proof of Theorem \ref{theo:main_theorem} }\label{sec:proof_maintheo} \label{sec:supp_convexprogram}
We start with rewriting \eqref{eq:newmodel_dualconst3} as follows
\begin{align}
\label{eq:newmodel_dualconst4}
&\max_{\mathcal{I}_j \in \{\pm1\}}\max_{\substack{i \in [P_1]\\ l \in [P_2]}}\max_{\substack{t_j\geq 0\\ \vec{1}^T\vec{t}\leq 1}}\max_{\substack{\|\weight_{1j}^{\prime}\|_2^2/w_{2j}^\prime \leq  t_j\\ \vec{1}^T\weight_2^\prime\leq 1\\ \weight_2^\prime \geq 0}} \, \dual^T \diag_{2l} \sum_{j=1}^{m_1}\mathcal{I}_j\diag_{1ij}\data \weight_{1j}^\prime \\
    &\text{ s.t. } (2\diag_{1ij} -\vec{I}_n)\data \weight_{1j}^\prime\geq 0,\,\forall j \in [m_1], \; (2\diag_{2l} -\vec{I}_n)\sum_{j=1}^{m_1} \mathcal{I}_j\diag_{1ij}\data \weight_{1j}^\prime \geq 0.\,  \nonumber
\end{align}
Then, the Lagrange function of \eqref{eq:newmodel_dualconst4} is as follows
\begin{align*}
    L(\weightmat_1^\prime,\boldsymbol{\alpha}_1,\boldsymbol{\alpha}_2)=\dual^T \diag_{2l} \sum_{j=1}^{m_1}\mathcal{I}_j\diag_{1ij}\data \weight_{1j}^\prime +\boldsymbol{\alpha}_2^T(2\diag_{2l} -\vec{I}_n) \sum_{j=1}^{m_1}\mathcal{I}_j\diag_{1ij}\data \weight_{1j}^\prime +\sum_{j=1}^{m_1} \boldsymbol{\alpha}_{1j}^T(2\diag_{1ij} -\vec{I}_n)\data \weight_{1j}^\prime
\end{align*}
where $\boldsymbol{\alpha}_2\geq0,\, \boldsymbol{\alpha}_{1j}\geq 0 ,\, \forall j \in [m_1]$. Thus, we have
%{\small
\begin{align}
    &\max_{\substack{i \in [P_1]\\ l \in [P_2]\\\mathcal{I}_j \in \{\pm1\}}}\max_{\substack{t_j\geq 0\\ \vec{1}^T\vec{t}\leq 1}}\max_{\substack{\|\weight_{1j}^{\prime}\|_2^2/w_{2j}^\prime \leq  t_j\\ \vec{1}^T\weight_2^\prime\leq 1\\ \weight_2^\prime \geq 0}} \min_{\substack{\boldsymbol{\alpha}_2 \geq 0\\ \boldsymbol{\alpha}_{1j}\geq 0}} \, \dual^T \diag_{2l} \sum_{j=1}^{m_1}\mathcal{I}_j\diag_{1ij}\data \weight_{1j}^\prime +\boldsymbol{\alpha}_2^T(2\diag_{2l} -\vec{I}_n) \sum_{j=1}^{m_1}\mathcal{I}_j\diag_{1ij}\data \weight_{1j}^\prime +\sum_{j=1}^{m_1} \boldsymbol{\alpha}_{1j}^T(2\diag_{1ij} -\vec{I}_n)\data \weight_{1j}^\prime \nonumber\\
      &=\max_{\substack{i \in [P_1]\\ l \in [P_2]\\\mathcal{I}_j \in \{\pm1\}}}\min_{\substack{\boldsymbol{\alpha}_2 \geq 0\\ \boldsymbol{\alpha}_{1j}\geq 0}}\max_{\substack{t_j\geq 0\\ \vec{1}^T\vec{t}\leq 1}} \max_{\substack{\|\weight_{1j}^{\prime}\|_2^2/w_{2j}^\prime \leq  t_j\\ \vec{1}^T\weight_2^\prime\leq 1\\ \weight_2^\prime \geq 0}} \, \sum_{j=1}^{m_1}\left(\mathcal{I}_j\dual^T \diag_{2l} \diag_{1ij}\data \weight_{1j}^\prime +\mathcal{I}_j\boldsymbol{\alpha}_2^T(2\diag_{2l} -\vec{I}_n) \diag_{1ij}\data \weight_{1j}^\prime + \boldsymbol{\alpha}_{1j}^T(2\diag_{1ij} -\vec{I}_n)\data \weight_{1j}^\prime\right) \nonumber \\
      &=\max_{\substack{i \in [P_1]\\ l \in [P_2]\\\mathcal{I}_j \in \{\pm1\}}} \min_{\substack{\boldsymbol{\alpha}_2 \geq 0\\ \boldsymbol{\alpha}_{1j}\geq 0}}\max_{\substack{t_j\geq 0\\ \vec{1}^T\vec{t}\leq 1}}\max_{\substack{ \vec{1}^T\weight_2^\prime\leq 1\\ \weight_2^\prime \geq 0}} \, \sum_{j=1}^{m_1}\left \| \mathcal{I}_j\data^T\diag_{2l} \diag_{1ij}\dual  +\mathcal{I}_j\data^T(2\diag_{2l} -\vec{I}_n) \diag_{1ij}\boldsymbol{\alpha}_2 + \data^T (2\diag_{1ij} -\vec{I}_n)\boldsymbol{\alpha}_{1j} \right\|_2 \sqrt{w_{2j}^\prime t_j} \nonumber \\
    &=\max_{\substack{i \in [P_1]\\ l \in [P_2]\\\mathcal{I}_j \in \{\pm1\}}} \min_{\boldsymbol{\alpha}_2,\boldsymbol{\alpha}_{1j}\geq 0} \max_{\substack{t_j\geq 0\\ \vec{1}^T\vec{t}\leq 1}} \, \left(\sum_{j=1}^{m_1}\left \| \mathcal{I}_j\data^T\diag_{2l} \diag_{1ij}\dual  +\mathcal{I}_j\data^T(2\diag_{2l} -\vec{I}_n) \diag_{1ij}\boldsymbol{\alpha}_2 + \data^T (2\diag_{1ij} -\vec{I}_n)\boldsymbol{\alpha}_{1j} \right\|_2^2 t_j\right)^{\frac{1}{2}} \nonumber \\
    &=\max_{\substack{i \in [P_1]\\ l \in [P_2]\\\mathcal{I}_{j_m}\in \{\pm1\}}} \min_{\substack{\boldsymbol{\alpha}_2 \geq 0\\ \boldsymbol{\alpha}_{1j_m}\geq 0}}\, \left \| \mathcal{I}_{j_m}\data^T\diag_{2l} \diag_{1ij_m}\dual  +\mathcal{I}_{j_m}\data^T(2\diag_{2l} -\vec{I}_n) \diag_{1ij_m}\boldsymbol{\alpha}_2 + \data^T (2\diag(S_{1{j_m}}) -\vec{I}_n)\boldsymbol{\alpha}_{1j_m}\right\|_2 ,
    %   &=\max_{\substack{t_j\geq 0\\ \|\vec{t}\|_2=s}}\min_{\boldsymbol{\alpha},\boldsymbol{\beta}_j\geq 0}  \, \|\vec{C}(\boldsymbol{\alpha},\boldsymbol{\beta})\|_F
     \label{eq:newmodel_dualconst5}
\end{align}%}
where $j_m$ denotes the index with the maximum norm. Note that we change the order of max-min for the first equality in \eqref{eq:newmodel_dualconst5} since the problem in \eqref{eq:newmodel_dualconst4} is convex and there exists a strictly feasible point, therefore strong duality holds, given fixed diagonal matrices $\{\diag_{1ij}\}_{j=1}^{m_1} $, $\diag_{2l}$ and a fixed set of signs $\{\mathcal{I}_j\}_{j=1}^{m_1}$.%Note that we remove the subscript $k$ for the rest of our analysis.

We first enumerate all hyperplane arrangements and signs and index them in an arbitrary order, which are denoted as $\diag_{1ij}$ and $\diag_{2l}$, where $i \in [P_1]$, $l \in [P_2]$, $P_1=|\mathcal{S}_1|$, and $P_2=| \mathcal{S}_2|$. %We remark that each arrangement for the first hidden layer $\diag_{1i}$ is associated with a sign pattern $\mathcal{I}_i$, thus, we include both $\diag_{1i}$ with $\mathcal{I}_i=+1$ and $\diag_{1i}$ with $\mathcal{I}_i=-1$ in the set $P_1$. 
Then we have
\begin{align*}
    \eqref{eq:newmodel_dualconst1} \iff & \max_{\substack{i \in [P_1]\\ l \in [P_2]\\\mathcal{I}_j \in \{\pm1\}}}\min_{\substack{\boldsymbol{\alpha}_2 \geq 0\\ \boldsymbol{\alpha}_{1j}\geq 0}}\max_{j \in [m_1]}  \left\| \mathcal{I}_{j}\data^T\diag_{2l} \diag_{1ij}\dual  +\mathcal{I}_{j}\data^T(2\diag_{2l} -\vec{I}_n) \diag_{1ij}\boldsymbol{\alpha}_2 + \data^T (2\diag_{1ij} -\vec{I}_n)\boldsymbol{\alpha}_{1j}\right\|_2  \leq \beta,\,
    \\\iff &\forall i \in [P_1], \forall j \in [m_1],\forall l \in [P_2],\forall \pm \;\exists \boldsymbol{\alpha}_{2il}^{\pm},\boldsymbol{\alpha}_{1ijl}^{\pm}\geq 0 \\
    &\hspace{3.cm}\text{ s.t. } \left\| \mathcal{I}_{ijl}^{\pm}\data^T\diag_{2l} \diag_{1ij}\dual  +\mathcal{I}_{ijl}^{\pm}\data^T(2\diag_{2l} -\vec{I}_n) \diag_{1ij}\boldsymbol{\alpha}_{2ijl}^{\pm} + \data^T (2\diag_{1ij} -\vec{I}_n)\boldsymbol{\alpha}_{1ijl}^{\pm} \right\|_2   \leq \beta,
\end{align*}
where we introduce the notation $\mathcal{I}_{ijl}^{\pm}=\pm1$ to enumerate all possible sign patterns. Therefore, the dual problem in \eqref{eq:newmodel_dual1} can also be written as 
\begin{align}
    \label{eq:newmodel_dual2}
      &  \max_{\substack{\dual\\\boldsymbol{\alpha}_{2ijl}^{\pm},\boldsymbol{\alpha}_{1ijl}^{\pm}\geq 0\\ \boldsymbol{\alpha}_{2ijl}^{\pm^\prime},\boldsymbol{\alpha}_{1ijl}^{\pm^\prime}\geq 0}} -\frac{1}{2} \| \dual- \vec{y}\|_2^2 +\frac{1}{2}\|\vec{y}\|_2^2 \\ \nonumber 
       &\text{ s.t. } \left\| \mathcal{I}_{ijl}^{\pm}\data^T\diag_{2l} \diag_{1ij}\dual  +\mathcal{I}_{ijl}^{\pm}\data^T(2\diag_{2l} -\vec{I}_n) \diag_{1ij}\boldsymbol{\alpha}_{2ijl}^{\pm} + \data^T (2\diag_{1ij} -\vec{I}_n)\boldsymbol{\alpha}_{1ijl}^{\pm} \right\|_2   \leq \beta,\;  \forall i \in [P_1], \forall j \in [m_1],\forall l \in [P_2],\forall \pm \\ \nonumber
       & \left\| -\mathcal{I}_{ijl}^{\pm^\prime}\data^T\diag_{2l} \diag_{1ij}\dual  +\mathcal{I}_{ijl}^{\pm^\prime}\data^T(2\diag_{2l} -\vec{I}_n) \diag_{1ij}\boldsymbol{\alpha}_{2ijl}^{\pm^\prime} + \data^T (2\diag_{1ij} -\vec{I}_n)\boldsymbol{\alpha}_{1ijl}^{\pm^\prime} \right\|_2  \leq \beta,\;  \forall i \in [P_1], \forall j \in [m_1], \forall l \in [P_2], \forall \pm.
\end{align}
We note that the above problem is convex and strictly feasible for $\dual=\boldsymbol{\alpha}_{1ijl}^{\pm}=\boldsymbol{\alpha}_{1ijl}^{\pm^\prime}=\boldsymbol{\alpha}_{2il}^{\pm}=\boldsymbol{\alpha}_{2il}^{\pm^\prime}=\vec{0}$. Therefore, Slater's conditions and consequently strong duality holds \cite{boyd_convex}, and \eqref{eq:newmodel_dual2} can be written as
\begin{align}
    \label{eq:newmodel_dual3}
      &\min_{\gamma_{ijl}^{\pm},\gamma_{ijl}^{\pm^\prime}\geq0} \max_{\substack{\dual\\\boldsymbol{\alpha}_{2ijl}^{\pm},\boldsymbol{\alpha}_{1ijl}^{\pm}\geq 0\\ \boldsymbol{\alpha}_{2ijl}^{\pm^\prime},\boldsymbol{\alpha}_{1ijl}^{\pm^\prime}\geq 0}} -\frac{1}{2} \| \dual- \vec{y}\|_2^2 +\frac{1}{2}\|\vec{y}\|_2^2 \\ \nonumber &+\sum_{+,-}\sum_{i=1}^{P_1}\sum_{j=1}^{m_1}\sum_{l=1}^{P_2}  \gamma_{ijl}^{\pm}\left(\beta- \left\| \mathcal{I}_{ijl}^{\pm}\data^T\diag_{2l} \diag_{1ij}\dual  +\mathcal{I}_{ijl}^{\pm}\data^T(2\diag_{2l} -\vec{I}_n) \diag_{1ij}\boldsymbol{\alpha}_{2ijl}^{\pm} + \data^T (2\diag_{1ij} -\vec{I}_n)\boldsymbol{\alpha}_{1ijl}^{\pm} \right\|_2  \right) \\ \nonumber
       &+\sum_{+,-}\sum_{i=1}^{P_1}\sum_{j=1}^{m_1}\sum_{l=1}^{P_2}\gamma_{ijl}^{\pm^\prime}\left(\beta - \left\| -\mathcal{I}_{ijl}^{\pm^\prime}\data^T\diag_{2l} \diag_{1ij}\dual  +\mathcal{I}_{ijl}^{\pm^\prime}\data^T(2\diag_{2l} -\vec{I}_n) \diag_{1ij}\boldsymbol{\alpha}_{2ijl}^{\pm^\prime} + \data^T (2\diag_{1ij} -\vec{I}_n)\boldsymbol{\alpha}_{1ijl}^{\pm^\prime} \right\|_2 \right),
\end{align}
where we change the order of max-min since strong duality holds. Next, we introduce variables $\vec{r}^{\pm}_{ijl},\vec{r}^{\pm^\prime}_{ijl} \in \mathbb{R}^{d}, \forall i \in [P_1], \forall j \in [m_1], \forall l \in [P_2], \forall \pm$ and represent the dual problem \eqref{eq:newmodel_dual3} as
\begin{align}
\label{eq:newmodel_dual4}
  & \min_{\gamma_{ijl}^{\pm},\gamma_{ijl}^{\pm^\prime}\geq0} \max_{\substack{\dual\\\boldsymbol{\alpha}_{2ijl}^{\pm},\boldsymbol{\alpha}_{1ijl}^{\pm}\geq 0\\ \boldsymbol{\alpha}_{2ijl}^{\pm^\prime},\boldsymbol{\alpha}_{1ijl}^{\pm^\prime}\geq 0}} \min_{\vec{r}^{\pm}_{ijl},\vec{r}_{ijl}^{\pm^\prime} \in \ball_2  } -\frac{1}{2} \| \dual- \vec{y}\|_2^2 +\|\vec{y}\|_2^2 \\ \nonumber &+\sum_{+,-}\sum_{i=1}^{P_1}\sum_{j=1}^{m_1}\sum_{l=1}^{P_2} \gamma_{ijl}^{\pm}\left(\beta+\vec{r}^{\pm^T}_{ijl}\left(  \mathcal{I}_{ijl}^{\pm}\data^T\diag_{2l} \diag_{1ij}\dual  +\mathcal{I}_{ijl}^{\pm}\data^T(2\diag_{2l} -\vec{I}_n) \diag_{1ij}\boldsymbol{\alpha}_{2ijl}^{\pm} + \data^T (2\diag_{1ij} -\vec{I}_n)\boldsymbol{\alpha}_{1ijl}^{\pm}  \right )\right) \\ \nonumber
   &+\sum_{+,-}\sum_{i=1}^{P_1}\sum_{j=1}^{m_1}\sum_{l=1}^{P_2}\gamma_{ijl}^{\pm^\prime}\left(\beta +{\vec{r}_{ijl}^{\pm^\prime}}^T\left(-  \mathcal{I}_{ijl}^{\pm^\prime}\data^T\diag_{2l} \diag_{1ij}\dual  +\mathcal{I}_{ijl}^{\pm^\prime}\data^T(2\diag_{2l} -\vec{I}_n) \diag_{1ij}\boldsymbol{\alpha}_{2ijl}^{\pm^\prime} + \data^T (2\diag_{1ij} -\vec{I}_n)\boldsymbol{\alpha}_{1ijl}^{\pm^\prime}  \right)  \right).
\end{align}
Now, we can change the order of max-min due to Sion's minimax theorem \cite{sion_minimax} and then compute the maximums with respect to $\dual,\boldsymbol{\alpha}_{2il}^{\pm},\boldsymbol{\alpha}_{1ijl}^{\pm},\boldsymbol{\alpha}_{2il}^{\pm^\prime},\boldsymbol{\alpha}_{1ijl}^{\pm^\prime}$
\begin{align}
\label{eq:newmodel_dual5}
  &\min_{\gamma_{ijl}^{\pm},\gamma_{ijl}^{\pm^\prime}\geq0}  \min_{\vec{r}^{\pm}_{ijl},\vec{r}^{\pm^\prime}_{ijl} \in \ball_2 } \frac{1}{2}\left\|\sum_{+,-}\sum_{j=1}^{m_1}\sum_{l=1}^{P_2} \sum_{i=1}^{P_1} \diag_{2l}\diag_{1ij} \data \left(\mathcal{I}^{\pm^\prime}_{ijl}\gamma^{\pm^\prime}_{ijl}\vec{r}^{\pm^\prime}_{ijl}-\mathcal{I}^{\pm}_{ijl}\gamma_{ijl}^{\pm}\vec{r}^{\pm}_{ijl}\right)-\vec{y}   \right\|_2^2+\beta \sum_{+,-}\sum_{i=1}^{P_1}\sum_{j=1}^{m_1}\sum_{l=1}^{P_2}\left( \gamma_{ijl}^{\pm}+\gamma^{\pm^\prime}_{ijl}\right) \\ \nonumber 
  &\text{ s.t. } (2 \diag_{2l}-\vec{I}_n)\mathcal{I}_{ijl}^{\pm} \diag_{1ij} \data\vec{r}^{\pm}_{ijl}\geq 0,\;
   (2 \diag_{1ij}-\vec{I}_n) \data\vec{r}^{\pm}_{ijl}\geq 0,\, \forall i \in [P_1], \forall j \in [m_1], \forall l \in [P_2], \forall \pm \nonumber\\
  &(2 \diag_{2l}-\vec{I}_n) \mathcal{I}_{ijl}^{\pm^\prime} \diag_{1ij} \data\vec{r}_{ijl}^{\pm^\prime}\geq 0,\;(2 \diag_{1ij}-\vec{I}_n)\data \vec{r}^{\pm^\prime}_{ijl}\geq 0,\, \forall i \in [P_1], \forall j \in [m_1], \forall l \in [P_2], \forall \pm. \nonumber
\end{align}
Next, we apply a change of variables as $\weight_{ijl}^{\pm}=\mathcal{I}_{ijl}^{\pm}\gamma_{ijl}^{\pm}\vec{r}^{\pm}_{ijl}$ and $\weight^{\pm^\prime}_{ijl}=\mathcal{I}_{ijl}^{\pm^\prime}\gamma^{\pm^\prime}_{ijl}\vec{r}^{\pm^\prime}_{ijl}$, which yields
\begin{align}
\label{eq:newmodel_final_proof}
  &\min_{\vec{w}_{ijl}^{\pm},\vec{w}_{ijl}^{\pm^\prime}} \frac{1}{2}\left\|\sum_{+,-}\sum_{j=1}^{m_1}\sum_{l=1}^{P_2} \sum_{i=1}^{P_1}\diag_{2l}   \diag_{1ij} \data \left(\vec{w}_{ijl}^{\pm^\prime}-\vec{w}_{ijl}^{\pm}\right)-\vec{y}   \right\|_2^2+\beta \sum_{+,-}\sum_{i=1}^{P_1}\sum_{j=1}^{m_1}\sum_{l=1}^{P_2}  \left( \|\vec{w}_{ijl}^{\pm}\|_2+\|\vec{w}_{ijl}^{\pm^\prime}\|_2\right) \\ \nonumber 
   &\text{ s.t. } (2 \diag_{2l}-\vec{I}_n)  \diag_{1ij} \data\vec{w}_{ijl}^{\pm}\geq 0,\;
   (2 \diag_{1ij}-\vec{I}_n) \data\vec{w}_{ijl}^{+}\geq 0,\;(2 \diag_{1ij}-\vec{I}_n) \data\vec{w}_{ijl}^{-}\leq 0,\, \forall i \in [P_1], \forall j \in [m_1], \forall l \in [P_2], \forall \pm \nonumber\\
  &(2 \diag_{2l}-\vec{I}_n) \diag_{1ij} \data\vec{w}_{ijl}^{\pm^\prime}\geq 0,\;(2 \diag_{1ij}-\vec{I}_n)\data \vec{w}_{ijl}^{+^\prime}\geq 0,\;(2 \diag_{1ij}-\vec{I}_n)\data \vec{w}_{ijl}^{-^\prime}\leq 0,\, \forall i \in [P_1], \forall j \in [m_1], \forall l \in [P_2],\forall \pm, \nonumber
\end{align}
which is a finite dimensional convex problem with $4dm_1 P_1 P_2$ variables and $4n(m_1+1) P_1 P_2$ constraints.

\hfill \qed

%%%%%%%%%%%%%%%%%%%%%%%%%%%%%%% New section- Proposition proof %%%%%%%%%%%%%%%%%%%%%%%%%%%%%%%%%%%%%%%%%
\subsection{Proof of Proposition \ref{prop:weight_construction}}\label{sec:proof_mapping}
We can construct an optimal solution to the primal problem in \eqref{eq:newmodel_primal1} from the optimal solution to the convex program in \eqref{eq:newmodel_final}, i.e., denoted as $\{\weight_{ijl}^{\pm^*},\weight_{ijl}^{{\pm^\prime}^*}\}_{i,j,l,\pm}$, as follows
\begin{align*}
& \weightmat_{1k}^*=\begin{cases}\frac{1}{\sqrt{\sum_{j=1}^{m_1}\|\weight_{ijl}^{+^*}\|_2}} \begin{bmatrix} \frac{\weight_{i1l}^{+^*}}{\sqrt{\|\weight_{i1l}^{+^*}\|_2}} & \ldots & \frac{\weight_{im_1l}^{+^*}}{\sqrt{\|\weight_{im_1l}^{+^*}\|_2}}   \end{bmatrix}&\text{ if } 1 \leq k \leq P_1 P_2\\
\frac{1}{\sqrt{\sum_{j=1}^{m_1}\|\weight_{ijl}^{-^*}\|_2}} \begin{bmatrix} \frac{-\weight_{i1l}^{-^*}}{\sqrt{\|\weight_{i1l}^{-^*}\|_2}} & \ldots & \frac{-\weight_{im_1l}^{-^*}}{\sqrt{\|\weight_{im_1l}^{-^*}\|_2}}  \end{bmatrix}&\text{ if } P_1P_2+1 \leq k \leq 2P_1 P_2\\
\frac{1}{\sqrt{\sum_{j=1}^{m_1}\|\weight_{ijl}^{{+^\prime}^*}\|_2}} \begin{bmatrix} \frac{\weight_{i1l}^{{+^\prime}^*}}{\sqrt{\|\weight_{i1l}^{{+^\prime}^*}\|_2}} & \ldots & \frac{\weight_{im_1l}^{{+^\prime}^*}}{\sqrt{\|\weight_{im_1l}^{{+^\prime}^*}\|_2}}  \end{bmatrix}&\text{ if } 2P_1 P_2+1 \leq k \leq 3P_1 P_2\\
\frac{1}{\sqrt{\sum_{j=1}^{m_1}\|\weight_{ijl}^{{-^\prime}^*}\|_2}} \begin{bmatrix}  \frac{-\weight_{i1l}^{{-^\prime}^*}}{\sqrt{\|\weight_{i1l}^{{-^\prime}^*}\|_2}} &\ldots & \frac{-\weight_{im_1l}^{{-^\prime}^*}}{\sqrt{\|\weight_{im_1l}^{{-^\prime}^*}\|_2}}  \end{bmatrix}&\text{ if } 3P_1 P_2+1 \leq k \leq 4P_1 P_2
\end{cases}\\
& \weight_{2k}^*=\begin{cases}\begin{bmatrix} \sqrt{\|\weight_{i1l}^{+^*}\|_2} & \ldots & \sqrt{\|\weight_{im_1l}^{+^*}\|_2} \end{bmatrix}^T&\text{ if } 1 \leq k \leq P_1 P_2\\
\begin{bmatrix}  \sqrt{\|\weight_{i1l}^{-^*}\|_2} & \ldots & \sqrt{\|\weight_{im_1l}^{-^*}\|_2} \end{bmatrix}^T&\text{ if } P_1 P_2 +1 \leq k \leq 2P_1 P_2\\
 \begin{bmatrix} \sqrt{\|\weight_{i1l}^{{+^\prime}^*}\|_2} &\ldots&\sqrt{\|\weight_{im_1l}^{{+^\prime}^*\|_2}}   \end{bmatrix}^T&\text{ if } 2P_1 P_2+1 \leq k \leq 3P_1 P_2\\
  \begin{bmatrix}  \sqrt{\|\weight_{i1l}^{{-^\prime}^*}\|_2} & \ldots & \sqrt{\|\weight_{im_1l}^{{-^\prime}^*}\|_2}  \end{bmatrix}^T&\text{ if } 3P_1 P_2+1 \leq k \leq 4P_1 P_2
\end{cases}\\
& \weightscalar_{3k}^*=\begin{cases}-\sqrt{\sum_{j=1}^{m_1}\|\weight_{ijl}^{+^*}\|_2}&\text{ if } 1 \leq k \leq P_1 P_2\\
-\sqrt{\sum_{j=1}^{m_1}\|\weight_{ijl}^{-^*}\|_2}&\text{ if } P_1P_2+1 \leq k \leq 2P_1 P_2\\
 \sqrt{\sum_{j=1}^{m_1}\|\weight_{ijl}^{{+^\prime}^*}\|_2}&\text{ if } 2P_1 P_2+1 \leq k \leq 3P_1 P_2\\
  \sqrt{\sum_{j=1}^{m_1}\|\weight_{ijl}^{{-^\prime}^*}\|_2}&\text{ if } 3P_1 P_2+1 \leq k \leq 4P_1 P_2
\end{cases},
\end{align*}
where
\begin{align*}
    (l,i) =\begin{cases}\left( \left\lfloor \frac{k-1}{P_1} \right\rfloor+1,  k- (l-1)P_1 \right) &\text{ if }1 \leq k \leq P_1 P_2 \\
    \left(\left\lfloor \frac{k-1-P_1P_2 }{P_1} \right\rfloor+1, k-P_1P_2 - (l-1)P_1 \right)&\text{ if }P_1 P_2+1 \leq k \leq 2P_1 P_2
    \\
    \left(\left\lfloor \frac{k-1-2P_1P_2 }{P_1} \right\rfloor+1, k-2P_1P_2 - (l-1)P_1 \right)&\text{ if }2P_1 P_2+1 \leq k \leq 3P_1 P_2
    \\
    \left(\left\lfloor \frac{k-1-3P_1P_2 }{P_1} \right\rfloor+1, k-3P_1P_2 - (l-1)P_1 \right)&\text{ if }3P_1 P_2+1 \leq k \leq 4P_1 P_2
    \end{cases}.
\end{align*}

% \begin{align*}
% &\forall k \in \{1,\ldots,P_1P_2\} \begin{cases}
%       \weight_{1kj}^*=\frac{1}{\sqrt{\sum_{j=1}^{m_1}\|\weight_{ijl}^*\|_2}} \frac{\weight_{ijl}^*}{\sqrt{\|\weight_{ijl}^*\|_2}} \\%\forall k \in \{1,2,\ldots,P_2\} \text{ where } j=k 
%       \weightscalar_{2kj}^*= \mathcal{I}_{ijl}^*\sqrt{\|\weight_{ijl}^*\|_2} \\
%       \weightscalar_{3k}^*=-\sqrt{\sum_{j=1}^{m_1}\|\weight_{ijl}^*\|_2} 
% \end{cases}, \text{ where } l=\left\lfloor \frac{k}{P_1} \right\rfloor+1 \text{ and } i= k- (l-1)P_1 \\
% &\forall k^\prime \in \{P_1P_2+1,\ldots,2P_1P_2\}  \begin{cases}
%       \weight_{1kj}^*=\frac{1}{\sqrt{\sum_{j=1}^{m_1}\|\weight_{ijl}^{\prime^*}\|_2}} \frac{\weight_{ijl}^{\prime^*}}{\sqrt{\|\weight_{ijl}^{\prime^*}\|_2}} \\%\forall k \in \{1,2,\ldots,P_2\} \text{ where } j=k 
%       \weightscalar_{2kj}^*= \mathcal{I}_{ijl}^{\prime^*}\sqrt{\|\weight_{ijl}^{\prime^*}\|_2} \\
%       \weightscalar_{3k}^*=\sqrt{\sum_{j=1}^{m_1}\|\weight_{ijl}^{\prime^*}\|_2} 
% \end{cases},\text{ where } l=\left\lfloor \frac{k^\prime}{P_1} \right\rfloor+1, i= k^\prime- (l-1)P_1, k^\prime=k-P_1P_2.
% \end{align*}
Therefore, we obtain an optimal solution to \eqref{eq:newmodel_primal1} as $\{\weightmat_{1k}^*, \weight_{2k}^*,\weightscalar_{3k}^*\}_{k=1}^{4P_1P_2}$, where $\weight_{1kj}^*$ and $\weightscalar_{2kj}^*$ are the columns and entries of $\weightmat_{1k}^* \in \mathbb{R}^{d \times  m_1}$ and $\weight_{2k}^* \in \mathbb{R}^{m_1}$, respectively. The optimality of these parameters can be verified as follows.

We first note that this set of parameters yields the same output with the convex program in \eqref{eq:newmodel_final}, i.e.,
\begin{align*}
    \sum_{k=1}^{4P_1P_2} \relu{\relu{\data \weightmat_{1k}^*}\weight_{2k}^*}\weightscalar_{3k}^*=\sum_{+,-}\sum_{j=1}^{m_1}\sum_{l=1}^{P_2} \sum_{i=1}^{P_1} \diag_{2l}  \diag_{1ij} \data \left(\vec{w}_{ijl}^{{\pm^\prime}^*}-\vec{w}_{ijl}^{\pm^*}\right).
\end{align*}
We also remark that these parameters are feasible for the original problem \eqref{eq:newmodel_primal1}, i.e., $\|\weightmat_{1k}^{*}\|_F^2=1,\, \forall k \in [4P_1P_2]$, and achieve the same regularization cost with \eqref{eq:newmodel_final}
\begin{align*}
   \frac{\beta}{2}\sum_{k=1}^{4P_1P_2} \left(\|\weight_{2k}^{*}\|_2^2+\weightscalar_{3k}^{*^2}\right)=\beta\sum_{+,-} \sum_{i=1}^{P_1}\sum_{j=1}^{m_1}\sum_{l=1}^{P_2}    \left( \|\vec{w}_{ijl}^{\pm^*}\|_2+\|\vec{w}_{ijl}^{{\pm^\prime}^*}\|_2\right)
\end{align*}
Since $\{\weightmat_{1k}^*, \weight_{2k}^*,\weightscalar_{3k}^*\}_{k=1}^{4P_1P_2}$ has the same output, therefore the same prediction error, and regularization cost with the optimal parameters of the convex program in \eqref{eq:newmodel_final}, this set of parameters also achieves the optimal objective value $P^*$, i.e.,
\begin{align*}
    P^*=\frac{1}{2}\left \| \sum_{k=1}^{4P_1P_2}\relu{\relu{\data \weightmat_{1k}^*}\weight_{2k}^* }\weightscalar_{3k}^* -\vec{y}\right\|_2^2+\frac{\beta}{2} \sum_{k=1}^{4P_1P_2}\left( \|\weight_{2k}^*\|_2^2+\weightscalar_{3k}^{*^2}\right).
\end{align*}
\hfill\qed

%%%%%%%%%%%%%%%%%%%%%%%%%%%%%%% New section- Generic loss proof %%%%%%%%%%%%%%%%%%%%%%%%%%%%%%%%%%%%%%%%%
\subsection{Proof for the dual problem in \eqref{eq:dual_overview}}\label{sec:dual_derivation}
The proof follows from classical Fenchel duality \cite{boyd_convex}. We first restate the primal problem after applying the rescaling in Lemma \ref{lemma:scaling_deep_overview}
\begin{align}\label{eq:overview_primal_supp}
   P^*= \min_{\hat{\vec{y}} \in \mathbb{R}^n,\theta \in \Theta_p} \mathcal{L}(\hat{\vec{y}},\vec{y}) + \beta \| \weight_L \|_1 \text{ s.t. } \hat{\vec{y}}=\sum_{k=1}^K\relu{\relu{\data \weightmat_{1k}} \ldots \weight_{(L-1)k} }\weightscalar_{Lk}.
\end{align}
Now, we first form the Lagrangian as 
\begin{align*}
    L(\dual,\hat{\vec{y}},\weight_L)=\mathcal{L}(\hat{\vec{y}},\vec{y})- \dual^T \hat{\vec{y}} +\dual^T \sum_{k=1}^K\relu{\relu{\data \weightmat_{1k}} \ldots \weight_{(L-1)k} }\weightscalar_{Lk}+ \beta \|\weight_L\|_1
\end{align*}
and then formulate the dual function as
\begin{align*}
    g(\dual)&= \min_{\hat{\vec{y}},\weight_L}L(\dual,\hat{\vec{y}},\weight_L)\\
    &=\min_{\hat{\vec{y}},\weight_L} \mathcal{L}(\hat{\vec{y}},\vec{y})- \dual^T \hat{\vec{y}} +\dual^T \sum_{k=1}^K\relu{\relu{\data \weightmat_{1k}} \ldots \weight_{(L-1)k} }\weightscalar_{Lk}+ \beta \|\weight_L\|_1 \\
    &=-\mathcal{L}^*(\dual)   \text{ s.t. } \left\vert \dual^T \relu{\relu{\data \weightmat_{1k}}\ldots \weight_{(L-1)k}} \right\vert\leq \beta,\, \forall k \in [K],
\end{align*}
where $\mathcal{L}^*$ is the Fenchel conjugate function defined as \cite{boyd_convex}
\begin{align*}
\mathcal{L}^*(\dual) := \max_{\vec{z}} \vec{z}^T \dual - \mathcal{L}(\vec{z},\vec{y})\,.
\end{align*}
Therefore, the dual of \eqref{eq:overview_primal_supp} with respect to $\weight_L$ and $\hat{\vec{y}}$ can be written as
\begin{align*}
   P^*= \min_{\theta  \in \Theta_p} \max_{\dual} g(\dual)=\min_{\theta  \in \Theta_p} \max_{\dual} -\mathcal{L}^*(\dual)   \text{ s.t. } \left\vert \dual^T \relu{\relu{\data \weightmat_{1k}}\ldots \weight_{(L-1)k}} \right\vert\leq \beta,\, \forall k \in [K].
\end{align*}

We now change the order of min-max to obtain the following lower bound
\begin{align*}
        P^*\geq D^* &=\max_{\vec{\dual}}\min_{\substack{\theta  \in \Theta_p}} - \mathcal{L}^*(\dual)   \text{ s.t. } \left\vert \dual^T \relu{\relu{\data \weightmat_{1k}}\ldots \weight_{(L-1)k}} \right\vert\leq \beta,\, \forall k \in [K]\\
         &=\max_{\vec{\dual}} - \mathcal{L}^*(\dual)   \text{ s.t. } \max_{\theta \in \Theta_p} \left\vert \dual^T \relu{\relu{\data \weightmat_{1}}\ldots \weight_{(L-1)}} \right\vert\leq \beta.
\end{align*}

\hfill\qed

%%%%%%%%%%%%%%%%%%%%%% new subsection- vector output%%%%%%%%%%%%%%%%%%%%%%%%%%%%%%%%%%
\subsection{Extension to vector outputs} \label{sec:supp_vectorout}
Here, we present the extensions of our approach to vector outputs, i.e., $\vec{Y} \in \mathbb{R}^{n\times C}$. The original training problem in this case is as follows
\begin{align*}
    P_v^*:=& \min_{\theta \in \Theta} \mathcal{L}\left( \sum_{k=1}^K f_{\theta,k}(\data),\vec{Y}\right)+ \frac{\beta}{2} \sum_{k=1}^K\sum_{l=L-1}^L \|\weightmat_{lk}\|_F^2 .
\end{align*}
Using the same scaling in Lemma \ref{lemma:scaling_deep_overview} and following the steps in the scalar output case yields the following dual problem
\begin{align*}
        &D_v^* :=\max_{\vec{\dualmat}}\min_{\substack{ \theta \in \Theta_p}} - \mathcal{L}^*(\dualmat)   \text{s.t. } \left \| \dualmat^T \relu{\relu{\data \weightmat_{1k}}\ldots \weight_{(L-1)k}} \right\|_2 \leq \beta,\, \forall k \in [K],
\end{align*}
where where $\mathcal{L}^*$ is the Fenchel conjugate function defined as \cite{boyd_convex}
\begin{align*}
\mathcal{L}^*(\dualmat) := \max_{\vec{Z}} \mathrm{trace}\left(\vec{Z}^T \dualmat\right) - \mathcal{L}(\vec{Z},\vec{Y})\,.
\end{align*}
The rest of the derivations directly follows the steps in Section \ref{sec:supp_convexprogram} and \cite{sahiner2021vectoroutput}.

\end{document}